\documentclass []{rusthesis3} 
\usepackage{amssymb}
\usepackage{amsthm}
\usepackage{cite}
\usepackage[cp1251]{inputenc}
\usepackage[english,russian]{babel}

\usepackage{algorithm}
\usepackage{algorithmic_r}

\usepackage{graphicx}
\newcommand{\ie}{т.\,е. \ }

\newcommand{\cal}{\mathcal}
\newtheorem{theorem}{Теорема}[section]
\newtheorem{lemma}{Лемма}[section]
\newtheorem{corollary}{Следствие}[section]
\newtheorem{prop}{Утверждение}[section]
\newtheorem{definition}{Определение}[section]
\newtheorem{exercise}{Упражнение}[section]
\newtheorem{note}{Замечание}[section]
\newcommand{\since}{так как }
\def\R{\mathsf{l\kern -.15em R}}
\def\Z{\bf Z}
\def\x{\bf x}
\def\y{\bf y}
\newcommand {\A}{\mbox{\bf A}}
\newcommand {\B}{\mbox{\bf B}}
\newcommand {\E}{\mbox{\bf E}}
\newcommand {\z}{\mbox{\bf z}}
\newcommand {\M}{{\bf M}}

\newcommand {\W}{{\bf W}}

\newcommand {\N}{{\bf N}}
\ssp
\newtheorem{myalgorithm}{Алгоритм}[chapter]
\newtheorem{proposition}{Утверждение}[chapter]

\begin{document}


\newcommand {\MyBox}{\Box}

\chapter*{}

\begin{center}
{\bf Конспект лекций по курсу <<Эволюционные алгоритмы>>}\\

\ \\

{\bf Еремеев А.В.}\\

\end{center}

{\bf Abstract.} This manuscript contains an outline of lectures on
"Evolutionary Algorithms"\ course read by the author in Omsk State
University n.a.~F.M.Dostoevsky. The course covers Canonic Genetic
Algorithm and various other genetic algorithms as well as
evolutioanry algorithms in general. Some facts, such as the
Rotation Property of crossover, the Schemata Theorem, GA
performance as a local search and "almost surely"\ convergence of
evolutionary algorithms are given with complete proofs. The text
is in Russian.\\

{\bf Введение}\\


Эволюционные алгоритмы (ЭА), к котоpым можно отнести эволюционные
стратегии, генетические алгоpитмы, эволюционное программирование,
беpут свое начало в pаботах А.Г. Ивахненко~\cite{Iv}, Л.А.
Растригина~\cite{Ras}, Дж.Холланда~\cite{Ho75},
И.Реченбеpга~\cite{Re73}, Л. Фогеля, А. Оуэнса, М.
Уолша~\cite{FOW69}, и дpугих автоpов, вышедших в 60-70-х годах
двадцатого века. Основная идея ЭА состоит в компьютеpном
моделиpовании пpоцесса эволюции. Пpи этом моделиpование
пpедназначено не для исследования биологических популяций, а для
pешения пpактических задач пpикладной
математики, в частности, задач оптимизации.\\

{Области применения ЭА:} Экономика, управление, инженерные задачи,
переработка информации,
космос, медицина и т.д.\\

{Научные дисциплины, на стыке которых возникла область
эволюционных
вычислений:}\\
1. Биология (генетика), \\
2. Прикладная математика (искусственный интеллект, методы оптимизации,
теория вероятности).\\

Постановка задачи безусловной оптимизации в общем виде:
\begin{equation}\label{discrprb}
\max\{f(x) : x \in X\},
\end{equation}
где $X$- пространство решений (мощности континуума, счетное или конечное).\\

\paragraph{Принцип работы ГА.} Один из типичных представителей эволюционных алгоритмов в
оптимизации -- генетический алгоритм (ГА). При запуске ГА
создается <<виpтуальная>> популяция особей (как пpавило достаточно
хpанить только их генотипы), каждая из которых пpедставляет
элемент пpостpанства pешений оптимизационной задачи. Здесь и далее
используются некоторые термины, заимствованные из
биологии~\cite{Reim}.

Пpиспособленность особей к условиям окpужающей сpеды выpажается
некотоpой монотонной функцией от значения целевой функции
задачи. Чем лучше pешение - тем выше приспособленность особей с
соответствующим генотипом. Популяция pазвивается за счет отбоpа
более пpигодных особей и пpименения к ним случайных опеpатоpов,
имитиpующих мутацию генов и рекомбинацию pодительских генотипов
(кроссинговер).

Отбоp может осуществляться по-pазному. Особенно распpостpаненными являются
опеpатоpы пpопоpциональной селекции (веpоятность выбоpа пpопоpциональна
пpигодности), сpезающей селекции (задается pавномеpным pаспpеделением
на множестве из $T$\% лучших генотипов популяции) и туpниpной селекции
($s$ особей извлекаются с помощью pавномеpного pаспpеделения, затем
беpется лучшая из них). Подробнее это будет обсуждаться в свое время.

ГА представляет собой универсальную схему для решения самых разных
задач. Достаточно определить представление решений в виде
генотипа, выбрать операторы селекции и кроссинговера, и алгоритм
можно применять. Более того, при достаточно общих предположениях
можно доказать, что с вероятностью единица алгоритм находит
оптимальное решение, если время его работы не ограничено.

\paragraph{Основные принципы эволюционных алгоритмов.}
В области эволюционных алгоритмов находит себе пpименение
бионический подход, состоящий в заимствовании пpинципов
организации систем из живой пpиpоды. В данном случае имеет место
использование принципа постепенных адаптивных преобразований в
пределах популяции или вида в ходе, так называемой, микpоэволюции.
Как показывают исследования акад.~Ю.П.~Алтухова и других авторов
(см.,~\cite{Altuhov},~\S~6.4), более <<масштабная>> межвидовая
изменчивость (макроэволюция) требует скачкообразных перестроек
генотипа и не может быть выведена непосредственно из постепенной
внутривидовой изменчивости.\footnote{С одной стороны,
микpоэволюция многокpатно наблюдалась и pаботоспособность этой
идеи не вызывает сомнения. С другой стороны, теория макpоэволюции
пока дает лишь правдоподобное объяснение устройства живых
организмов и палеонтологических находок. Стого говоpя, эта теоpия
является лишь шиpоко используемой гипотезой. Не совсем ясно, как
эту гипотезу доказывать или пpовеpять, т.к. для обpазования нового
вида тpебуется слишком пpодолжительное вpемя.} В связи с
недостаточной изученностью таких скачкообразных перестроек,
привлечение известной теории происхождения видов для обоснования
работоспособности ЭА представляется проблематичным и требуются
непосредственные исследования ЭА~(см.~\cite{RR02},~\S~1.3).

Среди специалистов по методам поиска распространено мнение, что
хотя эволюционные алгоритмы и подобны природным генетическим
<<механизмам>>, биологические принципы не следует рассматривать,
как ограничения при построении оптимизационных алгоритмов.
Л.А.~Растригин: <<... стремление к моделям механизмов
биологической эволюции не должно быть чрезмерным, т.к. они созданы
природой (и/или Творцом) для развития биологических существ.
Переносить их без поправок на развитие технических объектов было
бы серьезной методологической ошибкой.>>~\cite{Ra96}. Аналогичное
мнение высказывает C.R.~Reeves: <<Основные принципы отбора,
рекомбинации и мутации могут быть полезны при моделировании (хотя
и в довольно неточном и упрощенном виде) того, как система
приспосабливается к окружающей среде, но в случаях, когда интерес
представляет задача оптимизации, эти принципы сами по себе вряд ли
будут эффективны за исключением отдельных случаев.>>~\cite{Re97}.

{\bf Оценка вероятности возникновения требуемого белка в
результате мутации}:
\begin{itemize}
\item Число атомов на земле -- порядка $10^{50}$.

\item За один год популяция одной из наиболее быстро размножающейся бактерии {\em e.coli} проходит около 2000 поколений.

\item Возраст земли -- порядка $4\cdot10^9$ лет.
\end{itemize}

Тогда число испытаний в истории развитии популяции бактерий не превышает величины
порядка $10^{50} 10^{13} = 10^{63}$.

Длина последовательности в одной молекуле белка может быть около
150 аминокислот, одна аминокислота кодируется триплетом нуклеотидов, т.е. для ее кодировки необходимо 450 нуклеотидных 
позиций в молекуле ДНК. Было бы серьезным упрощением считать, что заданным требованиям удовлетворяет только
{\em одна} последовательность аминокислот (как, например, в~\cite{Dembski}).
Однако можно считать что примерно половина из нуклеотидных позиций (в нашем случае~225~шт.) имеют не больше двух <<подходящих>> нуклеотидов 
(ср., например,~\cite{web_logo}), а значит вероятность случайного получения требуемого нуклеотида в такой позиции -- не больше~0.5. 
Тогда вероятность получения подходящего белка целиком при однократном испытании (случайном выстраивании 450 нуклеотидов)
оценивается сверху величиной $2^{-225}\approx 10^{-156}$. Как было отмечно выше, общее число испытаний можно ограничить сверху числом $10^{63}$. Отсюда
получаем ничтожную оценку сверху для вероятности хотя бы
однократного получения подходящего белка за всю историю существования земли:~$10^{-93}$.
\pagebreak

\paragraph{Эвристики и метаэвристики.}
\mbox{ }\\

{\bf Эвристика} (от греч. еуриско - обнаруживаю) - метод решения, основанный
на неформальных, интуитивных соображениях, не гарантирующий получения
наилучшего решения. Попытки систематизировать эвристики принадлежат Р.Декарту,
Г.В.Лейбницу, Б. Больцано. (За основу взято определение из
Словаря по кибернетике.)

{\bf Метаэвристика} -- эвристика с универсальной схемой,
применимой для поиска приближенных решений различных
оптимизационных задач и представляющая собой итерационный процесс,
в котором многократно используются подчиненные эвристики,
учитывающие особенности задачи.


В метаэвристике могут использоваться различные принципы
исследования пространства решений и стратегии адаптации для учета
полученной информации.

На каждой итерации метаэвристики могут выполняться операции с единственным
текущим решением (или частичным решением), либо с набором (популяцией)
решений. Подчиненные эвристики могут быть процедурами высокого или низкого
уровня, простым локальным поиском, градиентным методом построения решения или
классическими оптимизационными процедурами.

Генетические алгоритмы (ГА) наряду с алгоритмами имитации отжига,
поиска с запретами и муравьиной колонии и др. относятся к классу
метаэвристик.

\chapter{Классический генетический алгоритм}

Генетический алгоритм (ГА) представляет собой эвристический
алгоритм оптимизации, в основу которого положены биологические
принципы естественного отбора и изменчивости. Процесс работы
алгоритма представляет собой последовательную смену поколений,
состоящих из фиксированного числа особей-точек пространства
решений,  причем особи с бо'льшим значением целевой функции (более
приспособленные) получают больше потомков в каждом следующем
поколении. Кроме того,  при формировании следующего поколения
часть потомков полностью идентична родителям, а часть изменяется
некоторым случайным образом в результате мутации и кроссинговера
(скрещивания).

При использовании генетического алгоритма для поиска в дискретном
пространстве~$X$, каждой строке из~$l$ символов некоторого
алфавита~$A$ должен быть сопоставлен элемент пространства~$X$ ,
т.е. определена функция $x : B \to X$, (называемая также {\em
схемой представления}), где $B=A^l$. Строки $\xi\in B$ принято
называть генотипами, а их образы $x(\xi)\in X$ -- фенотипами.

В классическом генетическом алгоритме~(КГА) используется двоичный
алфавит $A=\{0,1\}$.

Популяцией $\Pi=(\xi^1,\xi^2,...,\xi^{\lambda})$ численности $\lambda$ является
вектор пространства $B^{\lambda}$, координаты которого называются
генотипами особей (индивидов) данной популяции. Как правило,
нумерация особей популяции не имет значения. Численность популяции
${\lambda}$ фиксирована от начала работы алгоритма до  конца.
Предполагается, что ${\lambda}$ -- четное.

Целевая функция $f : X \to R$ исходной задачи заменяется в ГА на
функцию приспособленности генотипа\footnote{В англоязычной
литературе используется термин {\em fitness function.}} $\Phi(\xi)
= \phi(f(x(\xi)))$, где $\xi \in B$. Здесь $\phi: R \to R_+$ --
некоторая монотонно возрастающая функция. В биологической
интерпретации функция приспособленности отражает степень
приспособленности индивида с генотипом $\xi$ к условиям
"окружающей среды", заданным функцией $\Phi(\xi)$. При этом
максимумы целевой функции соответствуют наиболее приспособленным
генотипам для данной "окружающей среды". Простейшим примером
функции приспособленности  является сама целевая функция при
условии, что она неотрицательна.

Лучший из найденных генотипов к поколению~$t$ будем обозначать
через~$\tilde{\xi}^t$:
$$
\tilde{\xi}^t =\mbox{argmax}\{\Phi(\xi^{i,\tau}), i=1,...,{\lambda},
\tau=0,...,t\}.
$$

Приведем общую схему {\em генетического алгоритма с полной заменой
популяции}. Этой схеме соответствует, в частности, КГА.
Используемые здесь вероятностные операторы ${\mbox{Sel}: B^{\lambda} \to
\{1,\dots,{\lambda}\}}$, ${\mbox{Cross}: B
\times B \to B \times B}$ и ${\mbox{Mut}:B \to B}$ будут описаны ниже.\\

{\samepage {\bf Генетический алгоритм с полной заменой популяции}
\vspace{1em}\\
{\bf 0.} Положить $t:=0$.\\
{\bf 1.} Для $k$ от 1 до ${\lambda}$ выполнять: \\
{\bf 1.1.} Построить случайным образом генотип $\xi^{k,0}$.\\
{\bf 2.} Для $k$ от 1 до ${\lambda}/2$ выполнять шаги 2.1-2.3: \\
{\bf 2.1.} Селекция: выбрать генотипы $\xi:=\xi^{Sel(\Pi^t),t}$, $\eta:=\xi^{Sel(\Pi^t),t}$. \\
{\bf 2.2.} Скрещивание: построить $(\xi',\eta'):=\mbox{Cross}(\xi,\eta)$.\\
{\bf 2.3.} Мутация: положить $\xi^{2k-1,t+1} :=
\mbox{Mut}(\xi'), \ \ \xi^{2k,t+1}:=\mbox{Mut}(\eta')$.
\\
{\bf 3.} Положить $t:=t+1$.\\
{\bf 4.} Если $t\leq t_{max}$, то идти на шаг 2,
иначе -- на шаг 5.\\
{\bf 5.} Результатом работы КГА является лучшее из найденных
решений $x(\tilde{\xi}^{t_{max}})$.\\
}

Поясним приведенную схему. На этапе инициализации (шаги 0 и 1) формируется начальная
популяция~$\Pi^0$, элементы которой генерируются в соответствии
с равномерным распределением на множестве генотипов~$B$, т.е.
$P\{\xi^{i,0}_k=0\}=P\{\xi^{i,0}_k=1\}=1/2, \ i=1,\dots,{\lambda}, \
k=1,\dots,l$.

Вероятностный оператор селекции особей на пространстве популяций
${\mbox{Sel}(\Pi)}$ имеет то же значение, что и естественный
отбор в природе. Действие этого оператора состоит в выборе
номера родительской особи для построения очередного потомка.
Генотип~$\xi^{i,t}$ с номером~$i, \ i=1,\dots,{\lambda}$ из
популяции~$\Pi^t$ оказывается родительской особью при
формировании очередного генотипа~$\xi^{k,t+1}$
популяции~$\Pi^{t+1}$ с вероятностью
\begin{equation}\label{eq:Psel}
P_{\tt s}(i,\Pi^t)=\frac{\Phi(\xi^{i,t})}{\sum_{j=1}^{\lambda}
\Phi(\xi^{j,t})}.
\end{equation}
Если окажется, что $\sum_{j=1}^{\lambda} \Phi(\xi^{j,t})$, то есть все
генотипы имеют нулевую приспособленность, условимся выбирать номер
особи с равномерным распределением из 1,...,${\lambda}$.

В алгоритме не
исключается выбор~$\xi^{i,t}$ одновременно в качестве $\xi$ и
$\eta$ на шаге~2.1. Описанный оператор~Sel иногда также называют
селекцией методом рулетки~\cite{RutPilRut,Gold}. Предположим,
что колесо рулетки разбито на~${\lambda}$ секторов, причем сектор~$i$
соответствует особи~$i$ и имеет радианную меру~$2\pi P_{\tt
s}(i,\Pi^t)$. Тогда селекцию особи~$\xi^{i,t}$ можно
представлять, как выбор $i$-го сектора на колесе рулетки.

Данный оператор селекции также называется {\it пропорциональным} в
связи с тем, что при фиксированном составе популяции вероятность выбора особи
в качестве родителя пропорциональна ее приспособленности.

\paragraph{Процедуры кроссинговера и мутации.} Опишем двуместный оператор кроссинговера (скрещивания)
$\mbox{Cross}(\xi,\eta)$ и одноместный оператор мутации
$\mbox{Mut}(\xi)$, действие которых носит случайный характер.

Результат кроссинговера $(\xi',\eta')=\mbox{Cross}(\xi,\eta)$ с
вероятностью~$P_{\tt c}$ формируется в виде
$$
\xi'=(\xi_1, \xi_2,...,\xi_{\chi}, \eta_{\chi+1},...,\eta_l ),
$$
$$
\eta'=(\eta_1, \eta_2,..., \eta_{\chi}, \xi_{\chi+1},...,
\xi_l),
$$
где случайная координата скрещивания~$\chi$ выбрана c
равномерным распределением от 1 до~$l-1$. С вероятностью
${1-P_{\tt c}}$ оба генотипа сохраняются без изменений, т.е.
$\xi'=\xi, \ \eta'=\eta$. Влияние оператора кроссинговера
регулируется параметром~$P_{\tt c}$. Данный оператор принято
называть {\em одноточечным кроссинговером}.\footnote{В
англоязычной литературе используется термин {\em one-point
crossover.}}

Оператор мутации в каждой позиции генотипа с заданной
вероятностью~$P_{\tt m}$ изменяет ее содержимое. В противном
случае ген остается без изменений. Таким образом, мутация
элементов генотипа происходит по схеме Бернулли с вероятностью
успеха~$P_{\tt m}$.

Изменение вероятностей мутации и кроссинговера позволяет
регулировать работу KГА и настраивать его на конкретные задачи.
Увеличение вероятности мутации до~0.5 превращает КГА в простой
случайный перебор, имеющий весьма ограниченное применение
(см.~\cite{Ras}, \S~6.1). Уменьшение же~$P_{\tt m}$ до нуля
приводит к малому разнообразию генотипов в популяции и может
вызвать <<зацикливание>> КГА, когда на каждой итерации
генерируются лишь ранее встречавшиеся генотипы. Величины~$P_{\tt
c}$ и ${\lambda}$ также могут существенно влиять на скорость сходимости
популяции к решениям приемлемого качества (см.,
например,~\cite{Er2000, RutPilRut}). Настраиваемые параметры КГА
выбирают, как правило, в следующих диапазонах: $0 \le P_{\tt c}
\le 1, \ 10^{-3} \le P_{\tt m} \le 0.3, \ 30 \le {\lambda} \le 10000.$

В отличие от большинства представителей животного мира, особи
генетических алгоритмов имеют не двойной хромосомный набор
(диплоидный), а одинарный (гаплоидный), т.к. хранение
дублирующих друг друга генотипов, полученных потомком от обеих
родительских особей при решении оптимизационных задач не
целесообразно. Особи ГА сходны с такими организмами, как
мхи-гаметофиты или некоторые виды водорослей, которые имеют
одинарный набор хромосом в течение длительного этапа жизни.

\section{Способы кодировки решений, примеры использования КГА}

Рассмотрим задачу оптимизации с ограничениями:
\begin{equation}
\label{constrained}
\max\{F(x) : x \in D\subseteq X\},
\end{equation}
где $D$ -- область допустимых решений.

\subsection{Максимизация функции $f: \{a,a+1,...,b\}  \to {\bf N}$.}

Здесь $X=\Z, D=\{a,a+1,...,b\}$.
Воспользуемся бинарной кодировкой решений: $l=\lceil \log_2(b-a)
\rceil$,
\begin{equation} \label{std_bin}
 x(\xi)=a+\sum\limits_{j=0}^{l-1} \xi_{l-j} 2^j.
\end{equation}
При $x(\xi)\in \{a,a+1,...,b\}$ полагаем $\Phi(\xi)=f(x(\xi))$,
иначе: $\Phi(\xi)=0$.

\subsection{Задача о разрезе максимального веса.}

{\em Дан граф $G=(V,E), V=\{v_1,..., v_n\}$, каждому ребру
приписан вес $w:E \to R^+$. Найти разрез $\{U,U'\}: U\subseteq V,
U'=V\backslash U$, максимального веса}
$$
W(\{U,U'\})= \sum\limits_{e=(u,v)\in E: u \in U, v \in U'} w(e).
$$

Данная задача $NP$-трудна~\cite{GJ}.

Здесь $D=X=\{\{U,U'\} : U \subseteq V, U'=V \backslash U\},
x=\{U,U'\}, f(x)=W(x)$. Кодировка определяется так: $U(\xi)=\{v_j
\in V: \xi_j=1, j=1,...,n\}$, $l=n$, $x(\xi)=\{U(\xi),V\backslash
U(\xi)\}, \Phi(\xi)=f(x(\xi))$.

Очевидно, операторы $\mbox{Mut}$ и $\mbox{Cross}$ сохраняют допустимость
решений, поэтому они могут быть использованы непосредственно без каких-либо
усовершенствований.

Есть одна сложность -- вырожденность кодировки (иногда называют
"конкуренцией конвенций"), т.к. один и тот же разрез $\{U,U'\}$
может быть представлен двумя способами (либо 1 кодирует вершину,
лежащую в $U$, либо в $U'$). Такая неоднозначность может привести
к снижению эффективности работы ГА и бессмысленности скрещивания
особей, закодированных в разных "конвенциях".


\subsection{Применение КГА в непрерывной оптимизации,
способы кодировки решений, геометрический смысл кроссинговера }

\paragraph{Случай $D\subset {\bf R}^n$.} Если $D\subset X={\bf R}^n$ ограничено, его можно дискретизовать,
например, путем введения достаточно мелкой регулярной сетки. При
этом задача сводится к поиску оптимума на дискретной решетке в
некотором $n$-мерном параллелепипеде $\Omega \subset X$. Пусть
область $D$ погружена в $n$-мерный параллелепипед~$\Omega$:
$$
D \subseteq \Omega= \{x \in {\bf R}^n : a_1\leq x_1 \leq b_1,...,
a_n \leq x_n \leq b_n\}, \ d_i= b_i -a_i, \ i=1,...,n.
$$

Одним из наиболее "естественных"\ способов кодировки
представляется стандартная двоичная кодировка координат векторов в
строке генотипа.

Пример: (001 010 011 100) $\mathop{\longrightarrow}\limits^{x(\xi)}$ (1,2,3,4).

В общем виде:

\begin{equation}\label{bin1}
x(\xi)_i= a_i + \frac{d_i}{2^k-1}\sum\limits_{j=0}^{k-1} \xi_{ki-j}2^j,
\ \ i=1,...,n.
\end{equation}

Здесь предполагается, что на кодирование каждой координаты используется
$k$ бит, и подстрока $g^i=(\xi_{k(i-1)+1},..., \xi_{ki})$ кодирует $i$-ю
координату, $l=kn$.\\

\begin{prop} (О геометрическом смысле кроссинговера
(Р.Т.~Файзуллин~\cite{Faiz})\\
Пусть $i_2,i_3,...,i_n$ -- номера генов, с которых начинается
кодировка 2,3,...,$n$-й координат вектора фенотипа из ${\bf R}^n$.
Тогда если найдется такой $r$, что $\chi+1 = i_{r+1},\ 1 \le r <
n$, то результат скрещивания $(\xi',\eta')=Cross(\xi,\eta)$ может
быть получен некоторым поворотом $R_{\chi,\xi,\eta}$ родительских
фенотипов $x(\xi), x(\eta)\in {\bf R}^n$, оставляющим неподвижной
точку $x^0=\frac{x(\xi)+x(\eta)}{2}$. Т.е.
$x(\xi')=R_{\chi,\xi,\eta}(x(\xi))$, $\ \
x(\eta')=R_{\chi,\xi,\eta}(x(\eta))$. При этом середина отрезка,
соединяющего родительские фенотипы, $x^0=\frac{x(\xi)+x(\eta)}{2}$
остается неподвижной, т.е. $R_{\chi,\xi,\eta}(x^0)= x^0.$
\end{prop}

{\bf Доказательство.} Будем искать оператор~$R_{\chi,\xi,\eta}(x)$
в виде следующего аффинного преобразования:
\begin{equation} \label{matrA}
R_{\chi,\xi,\eta}(x)=A(x-x^0) + x^0.
\end{equation}

Случай 1. При $n-r$ четном предположим, что $A$ - диагональная
матрица с $r$ единицами в начале диагонали, далее заполненная
четным числом символов~$-1$. С помощью непосредственной проверки
легко убедиться, что отображением $A(x-x^0) + x^0$ с матрицей
указанного вида дает тот же результат, что и при действии
оператора кроссинговера. Действительно, для всех координат $j \le
r$ имеем
$$
(A(x(\xi)-x^0) +
x^0)_j=\frac{x(\xi)_j-x(\eta)_j}{2}+\frac{x(\xi)_j+x(\eta)_j}{2}=x(\xi)_j,
$$
$$
(A(x(\eta)-x^0) +
x^0)_j=\frac{-x(\xi)_j+x(\eta)_j}{2}+\frac{x(\xi)_j+x(\eta)_j}{2}=x(\eta)_j,
$$
а для всех $j$, таких что $r < j \le n$, имеем
$$
(A(x(\xi)-x^0) +
x^0)_j=-\frac{x(\xi)_j-x(\eta)_j}{2}+\frac{x(\xi)_j+x(\eta)_j}{2}=x(\eta)_j,
$$
$$
(A(x(\eta)-x^0) +
x^0)_j=-\frac{-x(\xi)_j+x(\eta)_j}{2}+\frac{x(\xi)_j+x(\eta)_j}{2}=x(\xi)_j.
$$
То есть, представление~(\ref{matrA}) корректно.

Рассмотрим 3-мерное подпространство, образованное координатами
$x_1, x_{j}, x_{j+1}$, при любом $j=r+1,r+3,...,n-1$. В этом
подпространстве действие кроссинговера описывается диагональной
подматрицей матрицы~$A$ с диагональю (1,-1,-1), задающей поворот
вокруг оси~$x_1$ на угол $\pi$. Преобразование матрицы~$A$ есть
композиция таких поворотов, следовательно, $A(x-x^0) + x^0$
является поворотом в $R^n$.

Случай 2. При $n-r$ нечетном рассмотрим матрицу~$A$ вида
\begin{tabular}{|cccc|}
    \   $E'$   \   & 0     & 0 &  0 . . . 0  \\
 0 . . . 0       & $\cos (-2\alpha)$ & $-\sin (-2\alpha)$&  0 . . . 0\\
 0 . . . 0       & $\sin (-2\alpha)$ & $\cos (-2\alpha)$ &  0 . . . 0\\
 0 . . . 0       & 0     & 0 & $-E''$ \\
\end{tabular} \ ,
где $\alpha=\mbox{arctg}
\frac{x(\eta)_{r+1}-x(\xi)_{r+1}}{x(\eta)_{r}-x(\xi)_{r}}$, а
единичные матрицы $E'$ и $E''$ имеют размерости $r-1$ и $n-r-1$,
соответственно. Непосредственная проверка, подобная сделанной в
случае~1, показывает что отображение $A(x-x^0) + x^0$ с матрицей
указанного вида дает тот же результат, что и оператор
кроссинговера.

Рассмотрим 2-мерное подпространство, образованное координатами
$x_{r}, x_{r+1}$: здесь матрица $A$ задает поворот на угол
$2\alpha$ вокруг начала координат. Во всех 3-мерных
подпространствах, образованных координатами $x _1, x_{j},
x_{j+1}$, при $j=r+2,r+4,...,n-1$ преобразование~$A$ является
поворотом вокруг оси~$x_1$ на угол $\pi$, как и в случае~1.
Cледовательно, $A(x-x^0) + x^0$ является поворотом в $R^n$. Q.E.D.

\subsection{Примеры применения КГА для задач с ограничениями}

\paragraph{Учет ограничений задачи.}
Существует несколько подходов к обработке точек из $X \backslash D$. \\

a) Использование штрафной функции. Существует следующий
простейший способ, который всегда применим: $f(x)=F(x)$ при $x
\in D$, иначе $f(x)=-M$, где $M$ -- достаточно большая константа.
Недостаток: все точки вне допустимой области одинаково плохи и ГА
не имеет информации о близости недопустимой точки к $D$. Во многих
задачах оптимизации поиск допустимого решения сам по себе
представляет достаточно сложную задачу (например, задача ЛП
сводится к поиску допустимой точки некоторой системы линейных
неравенств) и без такой дополнительной информации ГА может не
обнаружить ни одного допустимого решения.

Другие более эффективные способы использования штрафов (см.,
например, \cite{Karm,ST}) состоят в "градации недопустимости"\
решений. Например, если область $D$ задана системой неравенств
$D=\{x\in X={\bf R}^n : f_1(x) \leq 0,..., f_m(x)\leq 0\}$, в
качестве штрафной функции может быть использована $P(x)=
r\sum_{i=1}^m \max\{f_i(x),0\}$, где $r$ достаточно велико. Далее
полагают $f(x)=F(x)-P(x)$. Желательно, чтобы выполнялось условие
$F(x(\xi))-P(x(\xi))<F(x^*)-P(x^*)=F(x^*)$ для любого $\xi$
такого, что $x(\xi) \not\in D$.

Можно воспользоваться и классическим способом сведения задач
математического программирования к задачам безусловной
оптимизации. Для этого рассматривается возрастающая
последовательность (теоретически возрастающая до бесконечности)
$r_1, r_2,...$ Для каждого  $r_{\theta} ,\theta=1,2,...,\Theta$
решается задача (1) где $f(x)=F(x)-P(x)$ и $P(x)$ найдена при
$r=r_{\theta}$. Величина $\Theta$ задает число "больших"\ итераций
алгоритма и выбирается исходя из значимости выполенения
ограничений. Естественно при этом каждый новый запуск ГА
осуществлять не со случайной начальной популяции, а с последней
популяции предыдущего запуска (тогда каждое обновление
$r_{\theta}$ можно понимать как изменение "окружающей среды"\ с
точки зрения эволюции популяции).

b) Корректировка недопустимых решений -- с помощью методов
непрерывной оптимизации или каких-либо эвристик, стартуя с
недопустимого решения $x(\xi) \not\in D$, находят некоторое
допустимое решение $x' \in D$.

c) Выбор подходящей кодировки, при которой $x(B)\subseteq D$
(пример -- рассмотренная выше задача о разрезе максимального веса).\\

{\bf Пример 3.} Если $D$ -- множество точек шара, то недопустимые
решения могут проецироваться на границу шара и после этого
кодироваться как особи новой популяции.\\

{\bf Пример 4.} Задача целочисленного линейного
программирования~(ЦЛП)

Рассматривается задача ЦЛП следующего вида: найти

\begin{equation}
\label{ILP1}
F(x)=(c,x)\to \max \\
\end{equation} при условиях
\begin{equation}
\label{ILP2}
Ax \leq b,
x \geq 0,\\
\end{equation}
\begin{equation}
\label{ILP3}
x \in {\bf Z}^n.\\
\end{equation}

Здесь $A $ -- $ (m\times n) $ - матрица, $c=(c_1,...,c_n), \
b=(b_1,...,b_m)^T, \ x=(x_1,...,x_n)$. Далее предполагается, что
множество ${\cal M}$, определяемое системой
неравенств~(\ref{ILP2}), ограничено. Будем называть вектор
целочисленным, если все его компоненты целочисленны.

Общая схема ГА может быть легко адаптирована для задач целочисленного
линейного и нелинейного программирования. Ограничимся рассмотрением
ГА для задачи ЦЛП.

Многогранник допустимых решений погружается в $n$-мерный
параллелепипед $\Omega=\{x\in {\bf R}^n | 0 \leq x_j \leq d_j,
j=1,...,n \}$ с минимальным объемом. Границы параллелепипеда $d_j$
могут быть найдены решением $n$ соответствующих задач ЛП.

Минимальная длина битовой строки для кодировки координаты $j$
целочисленной точки из $\Omega$ имеет вид $k_j=\lceil \log_2
(d_j+1) \rceil$. При кодировке допустимым целочисленным точкам в
$\Omega$ сопоставляются элементы $B$, состоящие из $n$
последовательно записанных двоичных представлений координат:
\begin{equation}\label{bin2}
x(\xi)_j= \sum\limits_{i=0}^{k_j-1} 2^i \xi_{k_1+\dots+k_j-i}, \ \
j=1,...,n.
\end{equation}
Таким образом, пространство генотипов $B$ есть
$\{0,1\}^{k_1+...+k_n}$.

Оператор мутации вводится стандартным образом, а схема
кроссинговера может быть адаптирована для данной задачи с
помощью дополнительного условия: $\chi \in \{k_1, k_1+k_2,...,
k_1+...+k_{n-1}\}$. (Строго говоря, при такой модификации
алгоритм уже не является КГА.)

Функция приспособленности $\Phi(\xi)$ может быть определена
по-разному. Определим вспомогательную функцию
$$
f(x)= \left\{
\begin{array}{ccc}
F(x) & {\mbox{при}} & {s(x)=0} \vspace{2mm}
\\
{-Cs(x)} & {\mbox{при}} & {s(x)>0,}
\\
\vspace{-4mm}
&  &
\end{array}
\right.
$$

где $s(x)$ -- сумма нарушений системы ограничений (\ref{ILP2})
для точки $x$ и $C$ -- некоторая положительная константа,
величина которой достаточна, чтобы выполнялось $f(x')<f(x)$ для
всех $x, x' \in \Omega \cap {\bf Z}$, таких что $x\in {\cal M},
x' \not\in {\cal M}$.

Функция приспособленности может быть выбрана следующим
образом~\cite{Gold}:
$$
\Phi(\xi)=\frac{f(x(\xi))-f_{min}^t}{f_{avg}^t-f_{min}^t},
$$
где $f_{avg}^t, f_{min}^t$ -- среднее и минимальное значения
функции $f(x(\xi))$ на текущей популяции $\Pi^t$. Легко видеть,
что определенная таким образом функция $\Phi(\xi)$
неотрицательна и неубывает с ростом $f(x)$. Отметим, что такой
вариант функции приспособленности позволяет решать с помощью ГА
задачи без ограничения на знак целевой функции.

Решение задачи линейного программирования
(\ref{ILP1})-(\ref{ILP2}) назовем непрерывным оптимумом.

Модификация КГА: индивиды первого поколения порождаются с помощью $n$-мерного
нормального распределения с математическим ожиданием в точке
непрерывного оптимума и с последующим округлением дробных
координат.

Эксперимент показал, что если вероятность мутации достаточно
велика, дисперсия начального распределения может быть установлена
равной нулю. Кроме того, выяснилось, что как правило ГА
относительно быстро (по сравнению с точными алгоритмами, например,
Гомори) обнаруживает допустимые решения, близкие к оптимальному по
целевой функции, однако часто имеют место случаи преждевременной
сходимости\footnote{В англоязычной литературе исполльзуется термин
{\em premature convergence.}} ГА к некоторому приближенному
решению, в результате процесс поиска оптимума замедляется. Для
преодоления этого затруднения разработан точный гибридный
алгоритм~\cite{EKmos}, сочетающий ГА с перебором $L$-классов.\\

{\bf Пример 3. Простейшая задача размещения производства.}\\

Рассматриваается следующая задача оптимального размещения,
известная также как задача стандартизации~\cite{BGD}. Пусть
имеется возможность построить $m$ предприятий, каждое из которых
может обслуживать любого из $n$ клиентов. При этом открытие $i$-го
предприятия ($i=1,...,m$) стоит $c_i \geq 0$ единиц, а
обслуживание $j$-го клиента ($j=1,...,n$) на предприятии $i$
обходится в $C_{ij}\geq 0$ единиц стоимости. Задача состоит в
минимизации функционала
$$
F(z)=\sum\limits_{i=1}^m c_i z_i +
  \sum\limits_{j=1}^n \min \limits_{i: z_i=1} C_{ij},
$$
при условии, что
$$
\sum\limits_{i=1}^m z_i \ge 1,
$$
где $z_i\in \{0,1\}$. Решение $z^*$ представляет собой оптимальный
вектор - набор предприятий при поставленных условиях, причем по
вектору $z^*$ легко могут быть назначены и оптимальные
прикрепления клиентов к открытым предприятиям. Очевидно, что
единственным недопустимым решением является нулевой вектор
${z={\bf 0}}$.

Не теряя общности, можно предположить, что все $c_i>0$, т.к.
при $c_i=0$ предприятие $i$ можно включить в вектор $z$ решения задачи
в обязательном порядке -- полученный после этого план обслуживания будет
оптимален. При этом, если предприятие $i$ не оказалось назначенным ни
для одного клиента, его можно исключить из вектора решения.

Опишем схему КГА в применении к простейшей задаче размещения. В
качестве генотипа удобно взять вектор предприятий $z$.  В таком
случае пространство генотипов совпадает с пространством
фенотипов и отображение $x(\xi)$ -- тождественное. Для
определения функции приспособленности необходимо исходную задачу
минимизации с ограничениями свести к задаче максимизации без
ограничений: положим что при некотором достаточно малом $\varepsilon>0$
$$
\Phi(\xi)=
\left\{
\begin{array}{ccc}
\frac{1}{F(\xi)} & {\mbox{при}} & {\xi\neq 0} \vspace{2mm}
\\
\varepsilon & {\mbox{при}} & {\xi= 0.}
\\
\vspace{-4mm}
&  &
\end{array}
\right.
$$

Очевидно, что тогда любой допустимый вектор пространства решений
имеет б\'{о}льшую приспособленность
$\Phi(\xi)=1/f(x(\xi))=1/f(\xi)$, чем нулевой недопустимый
вектор.
Операторы мутации и кроссинговера полностью соответствует КГА.\\

Другой возможный подход: ГА с недвоичным представлением, где
$l=n$. $\xi_j \in \{1,...,m\}, \  j=1,...,n.$ Кодировка:
предприятие~$i$ обслуживает клиента~$j$ тогда и только тогда,
когда $\xi_j=i$, и если хотя-бы один клиент обслуживается
предприятием~$i$, то полагаем $z_{i}=1$.
Данная модификация генетического алгоритма уже не укладывается в схему КГА.\\

%

\section{ТЕОРЕМА О СХЕМАХ}

Схемой $H$ с $K$ фиксированными позициями будем называть множество
генотипов
$$
H=\{ \xi \in B | \xi_{j_1}=h_1, \xi_{j_2}=h_2,..., \xi_{j_K}=h_K\}, \ \
\mbox{ где } j_1<j_2, \ j_2<j_3,...,j_{K-1}< j_K.
$$
Число $K$ принято называть порядком схемы. Длиной $\delta(H)$
схемы $H$ будем  считать  расстояние  между крайними
фиксированными позициями,  т.е. $\delta(H)=j_K-j_1$. По
определению полагаем $\delta(B)=0$. Очевидно,  что при заданной
кодировке любая  точка пространства  состояний является  частным
случаем  схемы порядка $l$.

Ввиду того, что одним генотипом могут обладать несколько особей
популяции, далее удобно ввести обозначение $N(H,\Pi^t)$ для числа
представителей схемы $H$ в поколении $t$, аналогично
$N(\xi,\Pi^t)$ -- число представителей генотипа $\xi$ в поколении
$\Pi^t$. Введем специальное обозначение для среднего значения
функции приспособленности на особях схемы $H$ в поколении $t$:
$$
\Phi(H,\Pi^t)=\frac{\sum\limits_{i : \xi^{i,t}\in H}
\Phi(\xi^{i,t})} {N(H,\Pi^t)}.
$$

Рассмотрим оценку среднего числа представителей схем среди особей
нового поколения, иногда называемую теоремой о схемах или
фундаментальной теоремой генетических алгоритмов~\cite{Gold,
Ho75}.
\begin{theorem} Пусть $H$ -- схема порядка $K$ и величина $c$ такова, что
$\Phi(H,\Pi^t) \geq c\Phi(B,\Pi^t).$ Тогда в классическом
генетическом алгоритме
\begin{equation} \label{schemat}
E[N(H,\Pi^{t+1})] \geq c\cdot \left( 1-\frac{\delta(H) P_{\rm
c}}{l-1}\right) \cdot (1-P_{\rm m})^{K} N(H,\Pi^t).
\end{equation}
\end{theorem}

{\bf Доказательство.} Будем рассматривать очередную итерацию КГА
с номером~$t+1$ в вероятностном пространстве, определенном
описанной выше схемой КГА, его параметрами и совокупностью
особей популяции~$t$. Для начала рассмотрим вероятность того,
что выбранный при селекции генотип принадлежит~$H$. Она имеет
вид
\begin{equation}
\label{selinH} P\{\xi^{Sel(\Pi^t)} \in H \} = \sum_{i : \xi^{it}
\in H} \frac{\Phi(\xi^{it})}{\sum\limits_{j=1}^{\lambda} \Phi(\xi^{jt})}=
\frac{\Phi(H,\Pi^t)N(H,\Pi^t)}{\Phi(B,\Pi^t)\lambda} \geq c
\frac{N(H,\Pi^t)}{\lambda}.
\end{equation}

Кроме оператора селекции необходимо учитывать также действие
мутации и кроссинговера, которые могут разрушать, а могут и
создавать особи схемы $H$. Для получения искомой нижней оценки
будем  рассматривать только  возможность разрушения элементов
схемы~$H$. Рассмотрим случайную величину $\zeta_i\in \{0,1\}$,
равную 1, если~$\xi^{i,t+1} \in H$, и 0 иначе. Если мы оценим
снизу математическое ожидание для всех $\zeta_i$, то это даст
возможность получить оценку снизу и на
величину~$E[N(H,\Pi^{t+1})]$, т.к.
\begin{equation}\label{sum}
E[N(H,\Pi^{t+1})] = \sum_{i=1}^{\lambda} E[\zeta_i].
\end{equation}

С этой целью введем вспомогательную случайную величину $\zeta'_i$
для каждого $i=1,...,\lambda$. Пусть $\zeta'_i$ равна единице, если при
построении~$\xi^{i,t+1}$ на этапе кроссинговера все гены с
номерами $j_1,j_2,...,j_K$ были скопированы из одной родительской
особи, принадлежащей~$H$, и кроме того, при мутации ни один из
этих генов не был изменен. В противном случае $\zeta'_i =0$.
Очевидно, $\zeta_i \geq \zeta'_i$ для всех $i$.

Заметим, что неравенство~(\ref{selinH}) дает оценку вероятности
того, что родительский генотип, откуда при кроссинговере была
скопирована позиция $j_1$, принадлежал $H$. Введем обозначение для
следующего события: $\vartheta = \{\chi<j_1 \mbox{ или }\ \chi
\geq j_K\}$. Вероятность того, что при мутации ни один бит,
отвечающий за принадлежность к схеме $H$, не изменит свое
значение, равна $(1-P_{\rm m})^{K}$, т.к. генные мутации
происходят побитно и независимо. Таким образом, ввиду

независимости события кроссинговера от всех других событий в ГА,
имеем
\begin{equation} \label{pzeta}
P\{\zeta'_j=1\} \ge c\frac{N(H,\Pi^t)}{\lambda} (1-P_{\rm
m})^{K}P\{\theta\}.
\end{equation}

Из определения кроссинговера вытекает, что
$P\{\vartheta\}=1-\frac{\delta(H)P_{\rm c}}{l-1}$,
следовательно,
\begin{equation} \label{ezeta_prime}
 E[\zeta'_i] = P\{\zeta'_i=1\} \ge c\frac{N(H,\Pi^t)}{\lambda}
(1-P_{\rm m})^{K} \left(1-\frac{\delta(H)P_{\rm c}}{l-1}\right).
\end{equation}
Далее, из того что для любого $i=1,...,\lambda$
выполняется $E[\zeta_i] \ge E[\zeta'_i]$, с учетом~(\ref{sum})
получаем (\ref{schemat}).
Q.E.D.\\

Таким образом, при выборе кодировки решений разработчик ГА должен
стремиться к тому, чтобы перспективные свойства решений были бы
представлены в генотипе в виде как можно более коротких участков
хромосм. В таком случае эти свойства будут проще обнаруживаться в
процессе работы ГА и решения с такими свойствами будут активно
исследоваться.\\


{\bf Пример 1.} Рассмотрим случай, когда число представителей
схемы~$H$, состоящей из близких к оптимуму генотипов,
увеличивается. Пусть используется стандартная двоичная кодировка
решений $x \in \{0,1,\dots,2^l\}$ для функции $f(x)\equiv x$ и
$\Phi(\xi)=\sum\limits_{j=1}^{l} \xi_{j} 2^{l-j}$, $l\ge 2$.
Очевидно, $x^*=2^l-1, \ \xi^*=(1,1,...,1)$. Если фиксировать $k$
первых единиц, $0<k<l$, то в случае, когда начальная популяция
содержит всевозможные генотипы по одному экземпляру, имеем
$$
\Phi(H,\Pi^0)= \frac{\min\limits_{\xi \in H} \Phi(\xi)+
\max\limits_{\xi \in H} \Phi(\xi)}{2}= \frac{\sum\limits_{j=1}^{k}
2^{l-j}+2^l -  1}{2}=
$$
$$
\frac{2^l\sum\limits_{j=0}^{k} 2^{-j} -  1}{2}=
\frac{2^l \cdot \frac{1-\left(\frac{1}{2}\right)^{k+1}}{1-\frac{1}{2}}-1}{2}.
$$
С другой стороны, $\Phi(B,\Pi^0)=\frac{2^l-1}{2}.$ Поэтому
получаем оценку снизу:
$$
\frac{\Phi(H,\Pi^0)}{\Phi(B,\Pi^0)}=
\frac{2^{l+1}\left(1-\frac{1}{2^{k+1}}\right)-1}{2^{l}-1} \geq
2\left(1-\frac{1}{2^{k+1}}\right)=2-\frac{1}{2^k},
$$
т.к. $2(1-\frac{1}{2^{k+1}})\ge 1$ и $(xa-1)/(a-1) \ge x$ при $a >
1, x \ge 1$.

 Таким образом, пусть $c=2-2^{-k}$ и
$P_{\rm c}=1, k=l/4$. Тогда теорема о схемах дает:
$$
E[N(H,\Pi^1)] \geq c \left(1-\frac{l/4}{l-1}\right)(1-P_{\rm
m})^{l/4} N(H,\Pi^0),
$$
и при больших~$l$ правая часть приближается к $2 \cdot
\frac{3}{4}(1-P_{\rm m})^{l/4},$ что при $P_{\rm
m}<1-\sqrt[l/4]{2/3}$ превышает 1, т.е. при достаточно малой
вероятности мутации имеет место рост числа представителей схемы
$H$ в среднем. Например, при $l=100$ получаем $P_{\rm
m}<0.00003$.\\

{\bf Пример 2.} Если же фиксировать $k$ последних единиц
(обозначим такую схему $H'$), то
$$
\Phi(H',\Pi^0)= \frac{\min\limits_{\xi \in H'} \Phi(\xi)+
\max\limits_{\xi \in H'} \Phi(\xi)}{2}=
\frac{2^k-1+2^l-1}{2}=\frac{2^k(1+2^{l-k})-2}{2},
$$
и, следовательно,
$$
\frac{\Phi(H',\Pi^0)}{\Phi(B,\Pi^0)}<
\frac{1+2^{l-k}}{2^{l-k}-2^{-k}}.
$$
при $l-k\to \infty$ правая часть стремится к 1, т.е., при больших
длинах $l$ и существенно меньших $k$ теорема о схемах не будет
гарантировать рост числа представителей~$H'$. Например, при
$k=l/4$ и $l=100$ получаем
$$
\frac{\Phi(H',\Pi^0)}{\Phi(B,\Pi^0)}<\frac{1+2^{60}}{2^{60}-1}<1.0000000000000000018.
$$
В произведении со множителями, отвечающими за мутацию и кроссинговер, результат, скорее всего, уже будет меньше~1.

\section{Анализ разнообразия популяции}

Пусть $q\in \{1,\ldots,l\}$ -- некоторая позиция в генотипе.
Обозначим $H_q=\{\xi: \xi_q=0\}$ и рассмотрим величину
$$
a_q= \frac{\Phi(H_q,\Pi^t)N(H_q,\Pi^t)}{\Phi(B,\Pi^t)\lambda} =
\frac{\sum_{i: \xi^i \in H_q} \Phi(\xi^{it})}
 {\sum_{k=1}^{\lambda} \Phi(\xi^{kt})}, \ \
$$
как характеристику "качества"\ нулевого значения гена в позиции
$q$ и его распространения в популяции $\Pi^t$.

Получим формулы, определяющие зависимость между вероятностью
вырождения гена в позиции~$q$ от параметров генетического
алгоритма. Заметим, что $a_q$ есть вероятность выбора особи,
принадлежащей~$H_q$ из популяции~$\Pi^t$ при пропорциональной
селекции.

\begin{theorem}\label{Malutina} \cite{Mal} Для любого значения
$q\in \{1,\ldots,l\}$ в КГА:

\begin{equation}\label{eqn:loose1}
P\{N(H_q, \Pi^{t+1})=\lambda\}=(a_q+(1-2a_q)P_{\rm m})^{\lambda},
\end{equation}
\begin{equation}\label{eqn:loose0}
P\{N(H_q, \Pi^{t+1})=0\}=(1-a_q+(2a_q-1)P_{\rm m})^{\lambda}.
\end{equation}
\end{theorem}
{\bf Доказательство.} Рассмотрим подробно только равенство
(\ref{eqn:loose1}), так как доказательство (\ref{eqn:loose0})
проводится аналогично. Рассчитаем вероятность того, что для
каждого $i=1,\dots,{\lambda}/2,$ имеет место $\xi^{2i,t+1}\in H_q$ и
$\xi^{2i-1,t+1}\in H_q$, т.е. пара передаваемых в популяцию
$\Pi^{t+1}$ особей будет иметь нулевое значение в позиции $q$.
Обозначим, соответственно, $\eta^{2i}$ и $\eta^{2i-1}$ генотипы
этих особей перед применением к ним оператора мутации. Тогда
\begin{equation}\label{eqn:aftermut}
 P\{\xi^{2i,t+1}_q =0, \xi^{2i-1,t+1}_q = 0 \}
 =
 P\{\eta^{2i}_q = \eta^{2i-1}_q = 0\}(1 - P_{\rm m})^2 +
 \end{equation}
 $$
 +
 2P\{\eta^{2i}_q=1, \ \eta^{2i-1}_q = 0\}(1 - P_{\rm m}) P_{\rm m}
  +
 P\{\eta^{2i}_q = \eta^{2i-1}_q = 1\} P_{\rm m}^2.
$$
Рассчитаем вероятность $P\{\eta^{2i}_q = 0, \eta^{2i-1}_q = 0\}$.
В результате скрещивания могут получиться две особи, у которых в
позиции $q$ находится значение 0, только в том случае, если у
обеих родительских особей в позиции $q$ находилось значение 0.
Вероятность выбора таких особей, с учетом вероятностного смысла
величин~$a_q$ и независимости селекции каждой особи, равна
$(a_q)^2$.

Аналогично находим вероятность того, что у одной из полученных в
результате скрещивания особей в позиции $q$ находится значение 1,
а у второй -- значение 0. $P\{\eta^{2i}_q=1, \ \eta^{2i-1}_q = 0\}
= a_q (1-a_q).$

Далее, $P\{\eta^{2i}_q = \eta^{2i-1}_q = 1\} = (1-a_q)^2,$
так как в результате скрещивания могут получиться две особи, у
которых в позиции~$q$ находится значение~1, только в том случае,
если это значение находилось в позиции~$q$ у обеих родительских
особей.

Подставив найденные выражения в формулу (\ref{eqn:aftermut}),
получим, что для каждого $i,$ $i = 1,\dots,{\lambda}/2$:
\begin{equation}\label{eqn:p00}
 P\{\xi^{2i,t+1}_q = 0, \xi^{2i-1,t+1}_q= 0 \}
 =
\end{equation}
$$
 (a_q)^2 (1 - P_{\rm m})^2 + 2 a_q(1 - a_q) P_{\rm m}(1 - P_{\rm m})
 +
 (1 - a_q)^2 P_{\rm m}^2 =
$$
$$
 =(a_q (1 - P_{\rm m}) + (1- a_q)P_{\rm m})^2
 =(a_q + (1 - 2 a_q) P_{\rm m})^2.
$$
Так как ${\lambda}/2$ пар особей в $\Pi^{t+1}$ генерируются независимо
одним и тем же способом, то $P\{N(H_{q},\Pi^t)={\lambda}\}$ равна
$$
 P\{\xi^{2i,t+1}_q= 0, \xi^{2i-1,t+1}_q = 0\ \forall\ i = 1,\dots,{\lambda}/2\}
 =(a_q + (1 - 2 a_q) P_{\rm m})^{\lambda}.
$$
Q.E.D.\\

Данная теорема и следствие позволяют сделать вывод о том, что
вероятность вырождения гена и, следовательно, степень разнообразия
популяции зависят только от вероятности мутации и размера
популяции генетического алгоритма и не зависят от вероятности
скрещивания. Таким образом проявляется важное свойство
рассматриваемых операторов кроссинговера, которые порождают новые
комбинации из уже имеющихся "блоков", но при этом никакие
элементарные "блоки"\ не теряются.

Из теоремы следует, что вероятность вырождения значения 0 или 1 в
гене $q$ равна
$$
p(q,P_{\rm m},{\lambda})=(a_q+(1-2a_q)P_{\rm
m})^{\lambda}+(1-a_q+(2a_q-1)P_{\rm m})^{\lambda}.
$$
Таким образом, при $0<a_q<1$ вероятность вырождения гена в любой
позиции уменьшается с увеличением размера популяции ${\lambda}$, а также с
приближением  $P_{\rm m}$ к 1/2, ибо условие $\frac{\partial
p(q,P_{\rm m},{\lambda})}{\partial P_{\rm m}}=0$ эквивалентно
$$
(a_q+(1-2a_q)P_{\rm
m})^{{\lambda}-1}=(1-a_q+(2a_q-1)P_{\rm m})^{{\lambda}-1},
$$
что означает $P_{\rm m}=1/2$, т.к. на концах интервала
$p(q,0,{\lambda})=p(q,1,{\lambda})\ge p(q,1/2,{\lambda})$. Таким образом, наибольшее разнообразие достигается при 
$P_{\rm m}=1/2$, что и следовало ожидать.

С использованием полученной формулы для вероятности $p(q,P_{\rm m},{\lambda})$ можно на каждом шаге КГА выбирать 
такую наименьшую приемлемую вероятность мутации, которая дает желаемое количество невырожденных генов. Например, в качестве требования невырожденности популяции может быть условие того, чтобы число невырожденных позиций на каждом поколении было не менее~1.\\

\section{Теоретические результаты о времени первого достижения оптимума для КГА}

В теории ЭА, как правило, рассматриваются вопросы трудоемкости отыскания наилучшего возможного генотипа в процессе работы некоторого ЭА на выбранном классе задач. При  этом в первую  очередь интересуются средним числом пробных решений до первого получения оптимального генотипа, в зависимости от размерности задачи. Исследуемые классы задач могут содержать примеры сколь угодно большой размерности, как это принято в теории вычислительной сложности~\cite{GJ}. Для анализа времени первого достижения оптимального генотипа, как правило, используются такие методы теории вероятностей, как цепи Маркова, мартингалы, случайные процессы со сносом, стохастическое доминирование и др. 

Рассмотрим КГА при $p_c = 0, \ p_m = \chi / n $ с постоянным значением параметра~$ \chi> \ln 2 $.
Как выясняется, этот алгоритм неэффективен даже на классе линейных
функций приспособленности. 

\begin{theorem}\label{theorem:linear_approximation} \cite{DEL19}
 Если $f(x)\equiv \sum_{i=1}^n a_i x_i$, то КГА при численности популяции~$\lambda\ge
 n^{2+\delta},$ и вероятности мутации $\chi/n$ при константных $\delta>0$, $\chi>\ln(2)$, получает оптимум функции~$f$ 
 с вероятностью не более~$\lambda e^{-dn^{\delta}}$
 за $e^{cn}$ поколений, где $c,d$ -- положительные константы.
 \end{theorem}

Аналогичный вывод был получен
из анализа КГА при $ p_c = 1 $ в случае функции OneMax \cite{Oliveto2014, Oliveto2015}.

\begin{theorem}\label{thm:GA-on-adf} \cite{DEL19}
Если $f(x)=\sum_{i=1}^n a_i x_i$, то КГА при $p_c=0$, $p_m=\chi/n,$ где 
$\chi=(1-c)/(n a_{\max}),$ ${a_{\max}:=\max_{i=1}^n a_i,}$ для любой положительной константы $c<1$, 
и размере популяции
$\lambda \ge d n^2 a_{\max}^2 \ln({n}{a_{\max}})$ при достаточно большой константе  $d>0$, имеет среднее время достижения оптимума
$O(n^2 + n\lambda\ln{\lambda})$.
\end{theorem}


Аналогичный результат получен для КГА с вероятностью мутации  $\chi/n,$ где $\chi$-константа, но при условии, что приспособленность имеет вид $f(x)= s^{\sum_{i=1}^n a_i x_i}$, и $s>e^{\chi}$.

\chapter{Модификации генетических алгоритмов}

\section{Операторы селекции}

Для сравнения различных операторов селекции, которые будут описаны
далее, необходимо выбрать некоторую характеристику, которая имеет
одинаковый смысл для всех из них. Возмем в качестве такой
характеристики число, сколько раз фиксированная особь отбирается в
качестве родительской из популяции $\Pi^t$ в процессе построения
очередного поколения.

В процессе работы ГА, как правило, происходит сближение
приспособленности особей в популяции. В результате операторы
пропорциональной селекции все меньше <<отличают>> (в смысле
вероятности селекции) наиболее приспособленных особей от
отстающих. Способ решения этой проблемы, предложенный
Д.Голдбергом, состоит в масштабировании приспособленности в
процессе работы ГА~\cite{Gold} (см. приведенный выше пример
применения ГА к задаче ЦЛП). Альтернативный подход состоит в
использовании {\em ранжирования особей}.

Для фиксированной популяции $\Pi$ биекция $r_{\Pi}:\{1,...,{\lambda}\} \to
\{1,...,{\lambda}\}$ называется ранжированием, если для всех $i,j \in
\{1,2,...,{\lambda}\}$ выполняется:
$$\Phi(\xi^i)>\Phi(\xi^j) \ \Rightarrow r_{\Pi}(i)>r_{\Pi}(j).$$
Значение $r_{\Pi}(i)$ называется рангом особи $\xi^i$ в популяции
$\Pi$.\\

{\bf Ранговая селекция (ranking selection).} Предложена в работе
Goldberg \& Deb~\cite{GD91}. Пусть функция $\alpha: \{1,\dots,{\lambda}\}
\to \R_+$, такая что $\sum_{r=1}^{\lambda} \alpha(r) =1$. Тогда $\alpha$
называется ранжирующей функцией.

При заданной ранжирующей функции оператор селекции с
распределением вероятностей
$$
P\{\mbox{выбрать особь с рангом $r$}\}=\alpha(r), \ \ r=1,\dots,{\lambda},
$$
называется ранжирующей селекцией. Такая селекция не теряет
<<чувствительности>> при сколь угодно малых различиях
приспособленности особей.

Частный случай, где
$$
\alpha(r)=\frac{\eta-1}{{\lambda}} \left(\frac{2(r-{\lambda})}{{\lambda}-1} +
\frac{\eta}{\eta-1} \right),
$$
при $\eta \in (1,2]$ называется линейным ранжированием. Легко
видеть, что условия ранжирующей функции выполняются.

Особь с рангом~${\lambda}$ имеет вероятность селекции, равную~$\eta/{\lambda}$, а
особь с рангом~1 -- вероятность~$(2-\eta)/{\lambda}$. Если положить
$\eta=2$, то особь с рангом~1 имеет нулевую вероятность селекции
(наибольшая дифференциация селективности по рангу). Если же
$\eta\to 1$, то распределение вероятностей селекции стремится к
равномерному.

Ранговая селекция при $\eta=2-2/({\lambda}+1)$ состоит в применении
оператора селекции КГА с подстановкой рангов особей~$r_{\Pi}(i)$
вместо значений их приспособленности.

{\bf Упражнение.} Найти функцию $\alpha(i)$ для пропорциональной селекции, где в качестве приспособленности используется ранг особи.

{\bf Турнирная селекция (tournament selection).} Оператор
турнирной селекции с размером турнира $s$ (или оператор
$s$-турнирной селекции) при построении очередного решения из
текущей популяции извлекает $s$ особей с равномерным
распределением и выбирает лучшую из них (точнее, особь с
наибольшим рангом).

Сравним среднее число повторений фиксированной особи $\xi^{it}$
при 2-турнирной селекции (обозначаем далее через $Z_T(i,\Pi^t)$) и
при пропорциональной селекции КГА с подстановкой рангов особей
вместо значений их приспособленности (обозначаем через
$Z_{RR}(i,\Pi^t)$). Заметим, что в турнирной селекции вероятность
выбора особи~$i$ с рангом $r=r_{\Pi^t}(i)$ есть
$$
p(r)=C_s^1\frac{1}{{\lambda}}\left(\frac{r-1}{{\lambda}}\right)^{s-1}+
C_s^2\left(\frac{1}{{\lambda}}\right)^2\left(\frac{r-1}{{\lambda}}\right)^{s-2}+...+
C_s^s\left(\frac{1}{{\lambda}}\right)^s=
$$
$$
=\left(\frac{1}{{\lambda}}+\frac{r-1}{{\lambda}}\right)^s-\left(\frac{r-1}{{\lambda}}\right)^s=
\left(\frac{r}{{\lambda}}\right)^s-\left(\frac{r-1}{{\lambda}}\right)^s.
$$
В частности, при $s=2$ имеем:
$$
E[Z_T(i,\Pi^t)]={\lambda}p(r)=\frac{r^2-r^2+2r-1}{{\lambda}}=\frac{2r-1}{{\lambda}}.
$$
Сравним эту величину с $E[Z_{RR}(i,\Pi^t)]$. Легко видеть, что при
использовании ранжирования в пропорциональной селекции
$$
P_{sel}(i,\Pi^t)=\frac{2r}{{\lambda}({\lambda}+1)}, \ \
E[Z_{RR}(i,\Pi^t)]=\frac{2r}{{\lambda}+1},
$$
что приближается к $E[Z_T(i,\Pi^t)]$ при больших $r,{\lambda}$.

По формуле дисперсии для схемы Бернулли нетрудно показать, что
$$
D[Z_{RR}(i,\Pi^t)]=\frac{2r({\lambda}^2 -2r+{\lambda})}{{\lambda}^3+2{\lambda}^2+{\lambda}},
$$
$$
D[Z_T(i,\Pi^t)]=\frac{(2r-1)({\lambda}^2-2r+1)}{{\lambda}^3},
$$
и при ${\lambda}\to\infty$ имеем
$\frac{D[Z_{RR}(i,\Pi^t)]}{D[Z_T(i,\Pi^t)]} \to \frac{2r}{2r-1}$,
следовательно, при больших значениях ${\lambda}$ и $r$ 2-турнирная
селекция становится близка к стандартной пропорциональной селекции
и по дисперсии.\footnote{Именно особи с большим рангом $r$
оказывают наибольший эффект при построении очередной популяции.}

{\bf Упражнение.} Найти функцию $\alpha(i)$ для турнирной селекции.

Наконец, $(\mu,\lambda)$-селекция, один из наиболее простых операторов селекции в ЭА, аналогичен массовому отбору в растениеводстве и животноводстве: из популяции численностью $\lambda$ отбираются $\mu$~особей с наибольшими значениями функции приспособленности и родительские генотипы равновероятно выбираются из них для скрещивания и получения потомства.

\subsection{Стратегии управления популяцией}

\indent\indent Обновление всей популяции на каждой итерации КГА
соответствует подходу, применяемому при имитационном моделировании
в популяционной генетике (см., например,~\cite{Altuhov}), однако,
для ускорения поиска генотипов с высокой приспособленностью общая
схема ГА зачастую модифицируется. Основная мотивация при этом
состоит в том, что в КГА даже генотип, существенно превышающий по
пригодности все прочие особи популяции, с большой вероятностью
будет исключен из рассмотрения уже на следующей итерации после его
появления. Однако в успешных приложениях ГА приспособленность
потомков, как правило, имеет положительную корреляцию с
приспособленностью родительских генотипов. В таких случаях
целесообразно сохранять наиболее пригодные особи в течение ряда
итераций ГА и генерировать с помощью кроссинговера и мутации
оставшуюся часть популяции. Рассмотрим некоторые известные схемы
управления популяцией, реализующие этот принцип.

{\bf Элитарная стратегия.} На кажой итерации ГА, во-первых,
строится очередная популяция~$\Pi^{t+1}$ по правилам КГА.
Во-вторых, если по приспособленности все генотипы новой популяции
уступают максимально приспособленной ({\em элитной})
особи~$\xi^t_{\tt e}$ из предыдущей популяции, то один из наименее
приспособленных генотипов в~$\Pi^{t+1}$ заменяется на~$\xi^t_{\tt
e}$.

Таким образом, если ГА применяется для решения задачи безусловной
оптимизации~(\ref{discrprb}), то последовательность значений
целевой функции элитных фенотипов~$f(x(\xi^1_{\tt e})),
f(x(\xi^2_{\tt e})), \dots$ будет неубывающей. Более того, как
показал Г.~Рудольф~\cite{Rud}, при $x(B)=X$ и $0<P_{\tt m}<1$,
дополнение КГА элитарной стратегией обеспечивает сходимость
последовательности ~$f(x(\xi^1_{\tt e})), f(x(\xi^2_{\tt e})),
\dots$ к оптимальному значению целевой функции
задачи~(\ref{discrprb}) почти наверное.

Вместо одной элитной особи в ГА может сохраняться некоторое
подмножество генотипов текущей популяции, имеющих высокую
приспособленность (см., например,~\cite{EKmos}). Такие стратегии
называются {\em частичной заменой популяции}. Следующая стратегия
может рассматриваться как предельный случай расширения множества
элитных особей.

{\bf Стационарная стратегия управления популяцией.} При этой
стратегии на каждой итерации ГА в популяцию добавляются два
генотипа, полученных применением операторов кроссинговера и
мутации. Каждая новая особь замещает некоторый <<неперспективный>>
генотип. При этом в качестве <<неперспективного>> может быть взят
генотип с наименьшей приспособленностью, или генотип, выбранный с
равномерным распределением среди имеющих приспособленность ниже
средней в текущей популяции. В некоторых вариантах ГА на выходе
кроссинговера имеется только один генотип -- тогда изменяется
только одна особь популяции.

Особенностью стационарной стратегии управления популяцией является
значительно более быстрое <<сужение>> области поиска, по сравнению
с КГА. В связи с этим, во многих реализациях стационарной
стратегии управления популяцией при совпадении новой особи с одной
из имеющихся в популяции, новая особь в популяцию не добавляется.

{\bf Элитная рекомбинация (elitist
recombination)~\cite{GoldTheir94}} (не путать с популяцией с
элитой): особи текущей популяции случайным образом переставляются
и последовательно выбираются пары родительских особей
$(\xi^{1t},\xi^{2t}), (\xi^{3t},\xi^{4t}),...$ для скрещивания.
Каждая пара потомков сравнивается с соответствующими родительскими
особями,
и лучшие две из четырех особей помещаются в новую популяцию.\\

\section{Операторы скрещивания и мутации}

\indent\indent Наряду с оператором одноточечного кроссинговера
КГА, в генетических алгоритмах используются и другие операторы
рекомбинации родительских генотипов. Общей чертой для всех из них
является, так называемое, свойство {\em передачи генов}: значение
для каждого гена потомка выбирается из значений соответствующих
генов одного или другого родителей.\footnote{Данное свойство в
работах N.~Radcliffe назавно {\em gene transmission} и является
частным случаем {\em allele transmission} (см.~\cite{R91}).}

В некоторых вариантах кроссинговера результатом является один
генотип (см., например,~\cite{Er2000}), однако, наиболее
распространены операторы с двумя выходными генотипами. Во втором
случае, с целью сохранения разнообразия популяции, стремятся
построить как можно более удаленные один от другого генотипы
потомков.

Пусть даны родительские генотипы $\xi$ и $\eta$, для которых
порождается пара генотипов потомков $\xi', \ \eta'$. Если в задаче
отсутствуют ограничения, или схема представления решений такова,
что~$x(B) \subseteq D$, тогда уместно использовать, так
называемую, {\em маску кроссинговера}. Под этим термином понимают
вспомогательную последовательность ${\bf m}=(m_1,\dots,m_l) \in B
$, по которой строятся генотипы потомков:
$$
\begin{array}{ll}

\xi'_i= \left\{
\begin{array}{ll}
\xi_i, & \mbox{ если } \ m_i=1\\
\eta_i, &  \mbox{ иначе,}
\end{array} \right. ;\ \

&

\eta'_i= \left\{
\begin{array}{ll}
\eta_i, & \mbox{ если } \ m_i=1\\
\xi_i, &  \mbox{ иначе,}
\end{array} \right.
\end{array}
$$
для $i=1,\dots,l$. Рассмотрим два примера использования маски кроссинговера.\\

{\bf Равномерный кроссинговер.} Данный оператор определяется
выбором маски кроссинговера с равномерным распределением на
множестве~$B$. При действии этого оператора $i$-ый ген,
$i=1,\dots,l$, копируется в генотип потомка из $i$-той позиции
генотипа одного или другого родителя с равными
вероятностями, независимо от выбора других генов.\\

{\bf $k$-точечный кроссинговер.} Данный оператор представляет
собой обобщение одноточечного кроссинговера. В строке генотипа
выбирается~$k$ различных координат скрещивания $0<\chi_1 < \chi_2
< \dots < \chi_k<l$ с равномерным распределением среди
всевозможных таких наборов. Обозначим $\chi_0=0$, тогда маска
кроссинговера определяется следующим образом:
$$
m_i=\left\{
\begin{array}{ll}
1, & \mbox{ если } \ \max\{j: \ \chi_j < i\} \ \mbox{-- четное число} \\
0, &  \mbox{ иначе}
\end{array} \right.
$$
для $i=1,\dots,l$. Данный оператор имеет то свойство, что при
задании четного числа точек скрещивания~$k$, первая и последняя
координаты одного родителя всегда переходят одному потомку.
Наоборот, при нечетном~$k$ эти координаты копируются в каждый из
генотипов потомков от разных родителей.

Каждый из описанных операторов кроссинговера может быть реализован
и в варианте с одним генотипом потомка (для этого достаточно
отбросить второй генотип). Далее будет рассмотрен оператор
кроссинговера, при котором естественным образом строится
только один генотип потомка.

{\bf Непредвзятые операторы мутации.} Обобщением рассмотренных ранее операторов мутации является класс {\em  непредвзятых
операторов} мутации, введенный в работе~\cite{LehreWitt}.\footnote{Иногда этот термин переводится как 
{\em несмещенный оператор} мутации.}
В случае, когда генотип
является битовой строкой, непредвзятость мутаций состоит в
том, что вероятность того или иного действия над каждым битом не зависит от его значения и позиции.
Условия непредвзятости мутации $y=\mbox{Mut}(x)$ можно формально представить следующим образом.
\begin{enumerate}
\item Для любых $x,y,z,\in \{0,1\}^{\ell}$
$$
P\{y \mid x\} = P\{y \oplus z \mid x \oplus z\}
$$
\item Для любых $x,y\in \{0,1\}^{\ell}$ и любой перестановки ${\sigma:\{1,\dots,\ell\}\to\{1,\dots,\ell\}}$
$$
P\{y \mid x\} = P\{\sigma_b(y) \mid \sigma_b(x)\},
$$
где $\sigma_b(\cdot)$ обозначает применение перестановки $\sigma$ ко всем позициям битовой строки, т.е.
$\sigma_b(x_1,\dots,x_{\ell})=x_{\sigma(1)},\dots,x_{\sigma(\ell)}.$ 
\end{enumerate}
Легко проверить, что этому определению удовлетворяют оператор побитовой мутации из КГА и оператор точечной мутации: равновероятно выбрать одну позицию в генотипе и инвертировать выбранный бит.

\subsection{Недвоичная кодировка: особенности мутации и скрещивания}

Примеры: балансировка ротора~\cite{Re97} и задача наименьшего
покрытия~\cite{Er2000,BeCh}.

%

\subsection{Кроссинговер с частичным отображением для задач на перестановках}

Рассмотрим некоторые операторы, применяемые в задаче коммивояжера
и других задачах, где допустимыми решениями являются перестановки.
При описании этих операторов под случайным выбором понимается
выбор с равномерным распределением среди всех возможных
вариантов.\\

Оператор кроссинговера с частичным отображением был предложен в
работе Д.~Голдберга и Р.~Лингле~\cite{GoldLin85} и кратко
обозначается PMX.\footnote{От английского {\em partially mapped
crossover}.} Рассмотрим действие кроссинговера PMX на
иллюстративном примере.

Пусть даны следующие родительские генотипы с координатами скрещивания
$\chi_1=3, \chi_2=7$:

\begin{center}
\vspace{0.5em}
\begin{tabular}{c c c | c c c c | c c }
$\xi$=( 1 & 2 & 3 & 4 & 5 & 6 & 7 & 8 & 9)\\
$\eta$=( 4 & 5 & 2 & 1 & 8 & 7 & 6 & 9 & 3).
\end{tabular}
\end{center}

Сначала выполняется обмен средними участками генотипов, прочие гены при
этом считаются неопределенными:
\begin{center}
\vspace{0.5em}
\begin{tabular}{c c c | c c c c | c c }
( x & x & x & 1 & 8 & 7 & 6 & x & x)\\
( x & x & x & 4 & 5 & 6 & 7 & x & x).
\end{tabular}
\end{center}

Далее, для каждого из неопределенных значений проверяется, можно ли в этой
позиции оставить прежнее значение. Например, в первом из указанных
генотипов нельзя оставить~1 на первой позиции, т.к.~1 уже зафиксирована в
четвертом гене, однако можно оставить на месте значения 2, 3 и 9.
Аналогично, во втором генотипе можно оставить на месте 2, 9 и 3:

\begin{center}
\vspace{0.5em}
\begin{tabular}{c c c | c c c c | c c }
( x & 2 & 3 & 1 & 8 & 7 & 6 & x & 9)\\
( x & x & 2 & 4 & 5 & 6 & 7 & 9 & 3).
\end{tabular}
\end{center}

Наконец, остальные значения заполняются такими же попарными обменами,
какими изменялись средние участки, только теперь обмен происходит не между
генотипами, а внутри каждого из них. В рассматриваемом примере 4 меняется
на 1, 5 -- на 8, 6 -- на 7, 7 -- на 6. Таким образом, результат
кроссинговера имеет вид:

\begin{center}
\vspace{0.5em}
\begin{tabular}{c c c | c c c c | c c }
$\xi'$= ( 4 & 2 & 3 & 1 & 8 & 7 & 6 & 5 & 9)\\
$\eta'$=( 1 & 8 & 2 & 4 & 5 & 6 & 7 & 9 & 3).
\end{tabular}
\end{center}



Формализуем оператор PMX-кроссинговера в общем случае. Пусть пара
перестановок $\xi'$ и $\eta'$ вычисляется по заданным родительским
перестановкам $\xi$  и $\eta$. В процедуре PMX-кроссинговера два
индекса~$\chi$ и $\theta$, где $\chi<\theta$, выбираются с
равномерным распределением и компоненты перестановок $\xi$ и
$\eta$ от~$\chi$ до~$\theta$ копируются в перестановки потомков
$\xi'$ и $\eta'$ с обменом: $\xi'_j:=\eta_j$ и $\eta'_j:=\xi_j$,
$j=\chi,…,\theta$. Далее, для каждого $j\not\in
\{\chi,...,\theta\}$ проверяется, можно ли в этой позиции оставить
значение из родительского решения, а именно, если $\xi_j$
отсутствует среди скопированных генов, то полагаем
$\xi'_j:=\xi_j$. Аналогично модифицируется~$\eta'$.

Остальные компоненты $j\not \in \{\chi,...,\theta\}$ заполняются с
помощью отображений~$M^1(\eta_j)\equiv \xi_j$  и
$M^2(\xi_j)\equiv\eta_j$. В каждом гене
$j\not\in\{\chi,...,\theta\}$ строки~$\xi'$ полагаем
$\xi'_j:=M^1(\xi_j)$, если значение $M^1(\xi_j)$ еще отсутствует
в~$\xi'$; иначе $\xi'_j:=M^1(M^1(\xi_j))$, если значение
$M^1(M^1(\xi_j))$ еще отсутствует в~$\xi'$, и так далее. Вторая
перестановка заполняется аналогично с использованием
отображения~$M^2$.


\subsection{Порядковый кроссинговер для задач на перестановках~\cite{Davis85}}

\begin{center}
\vspace{0.5em}
\begin{tabular}{c c | c c c c c c c c | c c c c c}
2 & 1       &       3         & 4               &      5          &  6 & 7 & \hspace{7em}            &  2 & 1       & 4        & 3 & 6        & 7 & 5 \\
  &         &                 &                 &                 &    &   & $\longrightarrow$       &    &         &          &   &          &   &   \\
4 & 3       &       6         & 2               &      7          &  1 & 5 &                         &  4 & 3       & 2        & 1 & 5        & 6 & 7 \\
  & $\chi$  &                 &                 &                 &    &   &                         &    & $\chi$  &          &   &          &   &   \\
\end{tabular}
\end{center}

Данный оператор сохраняет абсолютные позиции элементов,
заимствованных от одного родителя, и относительные позиции
элементов, заимствованных у другого.




Как показали эксперименты, порядковый кроссинговер, также как кроссинговер PMX, дает хорошие результаты
в задачах составления расписаний, где большое значение имеет 
то, какие работы выполняются раньше, а какие -- позже.
В то же время, для задачи
коммивояжера и задач составления расписаний, где важно учитывать длительность
переналадки с одной работы на другую, 
лучшие результаты дают операторы рекомбинации, основанные на
наследовании свойства смежности вершин (см., например,~\cite{CT98,EK16}).

\subsection{Мутация в задачах на перестановках}

{\bf Мутация обмена\footnote{В англоязычной литературе принят термин {\em
exchange mutation.}}} состоит в обмене пары генов из случайно выбранных
позиций в данной на вход перестановке.
С точки зрения локального поиска для задачи коммивояжера, при
действии этого оператора выполняется шаг в случайно выбранную
точку из окрестности {\em 2-city swap}~\cite{Kochet}.\\

{\bf Мутация сдвига\footnote{В англоязычной литературе принят термин {\em
shift mutation.}}} состоит в перемещении гена из случайно выбранной
позиции на случайное число позиций влево или вправо. Содержимое всех
промежуточных генов при этом сдвигается на
одну позицию.\\


{\bf Мутация <<2-замена>>} определяется наиболее просто в случае
задачи коммивояжера в терминах фенотипов, то есть обходов
графа~$G$. В обходе, заданном входным генотипом случайным образом
выбираются два несмежных ребра и заменяются двумя новыми ребрами,
которые в данном случае определяются однозначно. С точки зрения
локального поиска, действие данного оператора представляет собой
шаг в случайно выбранную точку из окрестности, определенной
относительно 2-замены~\cite{PapaStig}.

\begin{exercise}Описать алгоритм, осуществляющий мутацию
<<2-замена>> в указанной выше недвоичной кодировке решений задачи
коммивояжера.
\end{exercise}


\section{Задача оптимальной рекомбинации}

Пусть решается задача условной максимизации в пространстве
двоичных строк длины~$\ell=n$. Рассмотрим вычислительную сложность
задачи отыскания <<наилучшего>> генотипа, как результата
кроссинговера для заданной пары родительских генотипов при условии
выполнения свойства передачи генов.

С учетом свойства передачи генов сформулируем {\em задачу
оптимальной рекомбинации}: для произвольных заданных родительских
генотипов $p^1, p^2$, представляющих допустимые решения, требуется
найти представляющий допустимое решение генотип~$\xi$, такой что:

1) для каждого $j=1,\dots,n$ выполняется $\xi_j =p^1_j$ или $\xi_j
=p^2_j$;

2) $\xi$ имеет максимальное значение приспособленности среди всех
генотипов, удовлетворяющих условию 1) и кодирующих при этом допустимые решения.

Далее множество номеров координат, в которых родительские генотипы
различны, будем обозначать через $D(p^1,p^2)$.

В качестве примера эффективно разрешимой задачи оптимальной
рекомбинации рассмотрим следующую известную задачу из теории
графов. Пусть имеется граф $G=(V,E)$ с множеством вершин
$V=\{v_1,\dots,v_n\}$ и множеством ребер~$E$. Задача о наибольшем
независимом множестве состоит в отыскании такого подмножества
${S\subseteq V}$, что ни одно ребро ${e \in E}$ не инцидентно
сразу двум вершинам из~$S$ (т.е. $S$ -- независимое множество) и
мощность этого множества максимальна.

Естественным будет представление решений с помощью
вектора-индикатора из $\{0,1\}^n$, где $\xi_j=1$ тогда и только
тогда, когда вершина~$v_j$ принадлежит искомому подмножеству.
Пусть $\Phi(\xi)=|x(\xi)|$ для любого допустимого
решения~$x(\xi)$. Как замечено в работе Э.Балаша и
В.Нихауса~\cite{BN98}, при использовании данного представления
решений задача оптимальной рекомбинации разрешима за
полиномиальное время.

Для того чтобы в этом убедиться, рассмотрим произвольные
родительские независимые множества $S_1$ и $S_2$ и соответствующие
им генотипы $p_1$ и $p_2$. Исходя из свойства передачи генов,
решение-потомок~$S$ должно содержать все множество вершин
$L=S_1\cap S_2$, кроме того, в~$S$ не должно быть элементов
множества $V\setminus (S_1\cup S_2)$, а вершины с номерами из
множества $D(p^1,p^2)$ необходимо выбрать оптимальным образом.
Последнее требование формулируется, как задача о наибольшем
независимом множестве в подграфе, порожденном множеством вершин с
номерами из $D(p^1,p^2)$. Легко видеть, что данный подграф
является двудольным.

Для отыскания наибольшего независимого множества в двудольном
графе~$H=(V',E')$ можно воспользоваться тем фактом, что наибольшее
независимое множество всегда является дополнением наименьшего
вершинного покрытия~$C'$, то есть такого наименьшего по мощности
множества вершин, что каждое ребро инцидентно хотя бы одной из
них.

Задача о наименьшем вершинном покрытии двудольного
графа~$H=(V',E')$ эффективно разрешима с помощью алгоритма
построения минимального разреза во вспомогательном графе,
состоящем из графа~$H$ и дополнительных вершины-источника~$v_0$ и
вершины-стока~$v_{n+1}$. Источник~$v_0$ соединяется со всеми
вершинами одной доли, а сток~$v_{n+1}$ -- со всеми вершинами
другой доли. Ребрам из множества~$E'$ приписываются бесконечные
пропускные способности, а ребрам, инцидентным дополнительным
вершинам -- единичные пропускные способности. Наименьшее вершинное
покрытие~$C'$ формируется из вершин, инцидентных ребрам
минимального разреза.

Генотип~$\xi$, являющийся вектором-индикатором множества $L \cup
(V'\setminus C')$ представляет собой решение задачи оптимальной
рекомбинации для задачи о наибольшем независимом множестве.

Приведенный результат Балаша и Нихауса может быть сформулирован
как

\begin{theorem} (Балаш, Нихаус \cite{BN98}) Задача оптимальной
рекомбинации для задачи о независимом множестве разрешима за
полиномиальное время.
\end{theorem}

\begin{exercise}Показать, что для задачи о наименьшем
вершинном покрытии при тех же предположениях о способе
представления решений (т.е. $v_j \in C \Leftrightarrow \xi_j=1$)
задача оптимальной рекомбинации эффективно разрешима.
\end{exercise}

Рассмотренная здесь постановка задачи оптимальной рекомбинации
может быть модифицирована -- см., например,~\cite{BE08,Er2000}.
Выбор наиболее подходящей формулировки этой подзадачи и методов ее
решения делается на основе вычислительного эксперимента.

\section{Генетический алгоритм как метод локального поиска}

\subsection{Задачи комбинаторной оптимизации}

Пусть $\{0,1\}^*$ обозначает множество всевозможных строк из нулей
и единиц произвольной длины, а $\N$ -- множество натуральных
чисел. Для~$S\in \{0,1\}^*$ символом~$|S|$ обозначается длина
строки~$S$.

Далее величина~$a>0$ будет называться полиномиально ограниченной
относительно величины~$b>0$, если существует полином с
положительными коэффициентами относительно~$b$, ограничивающий
сверху значения~$a$.

Пусть $\R$ обозначает множество вещественных чисел.

\begin{definition}\label{def:NPO}
Задача комбинаторной оптимизации -- это тройка ${{\mathcal
P}=(\mbox{\rm Inst},\mbox{\rm Sol},f_I)}$, где $\mbox{\rm Inst}
\subseteq \{0,1\}^*$ называется множеством индивидуальных задач
из~${\mathcal P}$, и
выполнены следующие условия:\\

1. Существует детерминированная машина Тьюринга, распознающая
принадлежность строки исходных данных~$I$
множеству~$\mbox{\rm Inst}$ за время, полиномиально ограниченное относительно~$|I|$.\\

2. $\mbox{\rm Sol}(I)\subseteq \{0,1\}^{n(I)}$ -- множество
допустимых решений индивидуальной задачи $I \in \mbox{\rm Inst}$,
причем размерность пространства решений~$n(I) \le \mbox{\rm
poly}(|I|)$ для некоторого полинома~$\mbox{\rm poly}$.

3. Для $I \in \mbox{\rm Inst}$ за полиномиально ограниченное время
относительно~$|I|$ вычислима целевая функция $f_I: \mbox{\rm
Sol}(I) \to {\R}^+$, которую требуется максимизировать (если
${\mathcal P}$ -- задача максимизации) или минимизировать (если
${\mathcal P}$ -- задача минимизации).
\end{definition}

Если различные решения имеют разную длину записи, то $n(I)$ --
наибольшая длина допустимого решения задачи. Далее через~$f_I^*$
обозначается оптимальное решение индивидуальной задачи~$I$, \ie
$f_I^*=\max\{ f_I({\bf x}) : {\bf x} \in {\rm Sol}(I)\}$, если
${\mathcal P}$ -- задача максимизации, либо $f_I^*=\min\{f_I({\bf
x}) : {\bf x} \in {\rm Sol}(I)\}$, если ${\mathcal P}$ -- задача
минимизации.

Далее ГА рассматривается в предположении $B=\{0,1\}^{n(I)}$ и
представление решений совпадает с кодировкой решений задачи~$\Pi$,
а задача комбинаторной оптимизации имеет критерий <<на максимум>>.

Кроме того будем предполагать, что при ${\bf x}\in \mbox{Sol}$,
функция приспособленности имеет вид $\Phi({\bf x}) = f({\bf x})$.
Если же ${\bf x}\not \in \mbox{Sol}$, то функция
приспособленности~$\Phi({\bf x})$ принимает значение меньше, чем
на любом допустимом решении, что соответствует штрафу за нарушение
ограничений задачи.

\subsection{Задача поиска локального оптимума}

Пусть для всякого элемента ${\eta}\in \mbox{\rm Sol}(I)$
определена некоторая его окрестность ${\mathcal
N}_I({\eta})\subseteq \mbox{\rm Sol}(I)$. Совокупность
$\{{\mathcal N}_I({\eta}): {\eta} \in \mbox{\rm Sol}(I)\}$
называется {\em системой окрестностей}.

\begin{definition} Если для~${\bf x} \in \mbox{\rm Sol}(I)$ при
всяком ${\eta} \in {\mathcal N}_I({\bf x})$ выполняется
неравенство $f_I({\eta})\leq f_I({\bf x})$ в случае задачи
максимизации или $f_I({\eta})\geq f_I({\bf x})$ в случае задачи
минимизации, то решение~${\bf x}$ называется локальным оптимумом в
системе окрестностей~${\mathcal N}_I$.
\end{definition}

Глобальный оптимум, т.е. оптимальное решение задачи комбинаторной оптимизации является частным случаем локального оптимума.

Если ${\mathcal D}(\cdot,\cdot)$ -- метрика, заданная для всех
элементов ${\bf x},{\eta} \in \mbox{\rm Sol}(I)$, то ${\mathcal
N}_I({\bf x})=\{{\eta}: {\mathcal D}({\bf x},{\eta}) \leq k\}, \ \
{\bf x} \in \mbox{\rm Sol}(I)$ называется {\em системой
окрестностей радиуса~$k$, порожденной метрикой ${\mathcal
D}(\cdot,\cdot)$}.

Алгоритм локального поиска начинает свою работу с некоторого
допустимого решения. Далее, на каждой итерации алгоритма происходит
переход от текущего решения к новому допустимому решению в его
окрестности, имеющему лучшее значение целевой функции, чем текущее
решение. Процесс продолжается, пока не будет достигнут локальный
оптимум. Способ выбора нового решения в окрестности текущего
решения зависит от специфики конкретного алгоритма локального
поиска.

\subsection{Достижение локальных оптимумов генетическим алгоритмом}

Настоящий раздел посвящен изучению достаточных условий, при
которых генетический алгоритм с полной заменой популяции и
турнирной селекцией впервые посещает локальный оптимум в среднем
за время, близкое к трудоемкости локального поиска (точнее, превышающее ее не более, чем в $\log(n)$ раз). Ограничим рассмотрение задачами безусловной оптимизации
вида~(\ref{discrprb}).

Мотивацией исследования служит тот факт, что ГА зачастую относят к
классу методов локального поиска (см., например,~\cite{Kochet}),
поэтому представляет интерес детальное изучение случаев, когда
работоспособность ГА объясняется сходством его поведения с
локальным поиском.

Для простоты обозначений здесь предполагается двоичное
представление решений, совпадающее с кодировкой решений задачи
комбинаторной оптимизации, а <<генотип>> -- то же, что элемент
пространства решений~$\{0,1\}^{n(I)}$. В связи с этим для
обозначения генотипов, как правило, будут использоваться
символы~$\x$ или $\y$.

Исследуется ГА с полной заменой популяции и турнирной селекцией.
Для удобства анализа будем считать, что условие остановки ГА
никогда не выполняется.

Будем предполагать, что в результате кроссинговера с вероятностью
не менее некоторой константы~$\varepsilon$, $0<\varepsilon \le 1$,
образуются особи $({\xi}',{\eta}')=\mbox{Cross}({\xi},{\eta})$,
хотя бы одна из которых не уступает по приспособленности
родительским особям ${\xi},{\eta} \in B$, т.~е.
\begin{equation}\label{eps_cross}
{\bf P}\big\{\max\{\Phi({\xi}'),\Phi({\eta}')\} \ge
\max\{\Phi({\xi}),\Phi({\eta})\}\big\} \ge \varepsilon
\end{equation}
при любых ${\xi},{\eta} \in B$. Под <<констаной>> в настоящем
разделе понимается величина, не зависящая от индивидуальной
задачи.

Для одноточечного кроссинговера условие~(\ref{eps_cross})
выполняется c $\varepsilon = 1-P_{\rm c}$, если $P_{\rm c}<1$ --
константа, не зависящая от задачи. Условие~(\ref{eps_cross})
выполняется c $\varepsilon = 1$, если один из двух потомков --
решение задачи оптимальной рекомбинации родительских решений.

Пусть имеется задача комбинаторной оптимизации~${{\mathcal
P}=(\mbox{\rm Inst},\mbox{\rm Sol},f_I)}$ на максимум, причем
$\mbox{\rm Sol}(I)=\{0,1\}^{n(I)}$. Последнему условию
удовлетворяют многие задачи комбинаторной оптимизации, например,
задача максимальной выполнимости логической формулы~\cite{GJ},
разрез наибольшего веса~\cite{GJ}, спиновое стекло в модели
Изинга~\cite{Barahon} и др.

Пусть выбрана некоторая система окрестностей~$\{{\mathcal
N}({\xi}) \ | \ {\xi} \in {\rm Sol}(I)\}$. Обозначим через~$h$
число всех неоптимальных значений целевой функции~$f$, т.~е.
$h=|\{f({\xi}): {\xi} \in \mbox{\rm Sol} \}|-1$. Тогда, начиная с
любого допустимого решения, локальный поиск достигает локального оптимума не
более чем за~$h$ улучшающих целевую функцию итераций. Пусть~$L$
обозначает минимальную вероятность достижения решения в пределах
окрестности:
$$
L=\min_{{\xi} \in {\rm Sol}, \ {\xi}' \in {\mathcal  N}
({\xi})}{\bf P}\{\mbox{Mut}({\xi})={\xi}'\}.
$$
Чем выше величина~$L$, тем больше согласованность оператора
мутации с системой окрестностей. Численность популяции~${\lambda}$, размер
турнира~$s$ и величину~$L$ будем рассматривать как функции от
исходных данных задачи~$I$.

Пусть $e$ -- число Эйлера.\\

\begin{lemma}\label{lemma_lb}
$e^{-x} \ge 1-x.$
\end{lemma}
Действительно,
$$
e^{-x} = 1-x+x^2/2!-... \ge 1-x.
$$
\\

\begin{lemma}\label{lemma_ub}
$e^{-x} \le 1-x/e$ при $x\in [0,1]$.
\end{lemma}
Заметим, что $2(1-1/e)>1,$ а поэтому при $x\in [0,1]$ имеем $x \le
2(1-1/e)$, т.е. $1-x/2 \ge 1/e$ и $x-x^2/2 \ge x/e$. Далее,
$$
e^{-x} = 1-x+x^2/2!-... \le 1-x+x^2/2\le 1-x/e.
$$
\\

\begin{lemma}\label{lemma_cond}
Для любых событий $A_0,A_1,...,A_n$
$$
P\{A_0\& A_1\& ... \& A_n\} = P\{A_0\} \prod_{i=0}^{n-1}
P\{A_{i+1}|A_0 \& A_1\& ... \& A_i\}.
$$
\end{lemma}

Действительно, по определению условной вероятности,
$$
P\{A_0\& A_1\& ... \& A_n\} = P\{A_n|A_0\& A_1\& ... \&
A_{n-1}\}\cdot  P\{A_0\& A_1\& ... \& A_{n-1}\}= ...
$$
$$
= P\{A_0\} \prod_{i=0}^{n-1} P\{A_{i+1}|A_0 \& A_1\& ... \& A_i\}.
$$
\\

\begin{theorem}\label{th_GA_LS} \cite{Er_diser}
Если $s \ge r{\lambda}$, $r>0$, $h> 1$, $L>0$ и
\begin{equation}\label{N_condition}
{\lambda} \ge \frac{2(1+ \ln h)}{L\varepsilon(1-1/e^{2r})},
\end{equation}
то
\begin{enumerate}
\item GA посещает локальный оптимум к итерации~$h$ с вероятностью не менее $1/e$, и
\item локальный оптимум достигается не позднее, чем за~$eh$
итераций GA в среднем.
\end{enumerate}
\end{theorem}

{\bf Доказательство.} Пусть популяция~$\Pi^t$ еще не содержит локального оптимума и событие $E_k^{t+1}$, $k=1,\dots,{\lambda}/2$,
состоит в выполнении следующих трех условий:
\begin{enumerate}
\item из популяции~$\Pi^t$ при построении $k$-той пары потомков
следующего поколения выбирается решение~${\xi}^t_*$ наибольшей
приспособленности;
\item при построении $k$-той пары
потомков посредством кроссинговера, один из них имеет
приспособленность не менее $\Phi({\xi}^t_*)$ (пусть для
определенности это~${\xi}'$);
\item оператор мутации,
примененный к~${\xi}'$, осуществляет переход в наилучшее по
приспособленности решение в окрестности~${\mathcal  N}({\xi}')$,
т.~е. $\Phi(\mbox{Mut}({\xi}'))=\max_{{\eta} \in {\mathcal
N}({\xi}')} \Phi({\eta})$.
\end{enumerate}

Обозначим через~$p$ вероятность наступления хотя бы одного из
событий $E_k^{t+1}, \ k=1,\dots,{\lambda}/2$, при известной
популяции~$\Pi^t$. Найдем оценку $p_{LB} \le p$, не зависящую от
выбора~$\Pi^t$. Согласно схеме GA, ${\bf P}\{E_1^{t+1}\}=\dots
={\bf P}\{E_{{\lambda}/2}^{t+1}\}$. Обозначим эту вероятность через~$q$.
Ввиду независимости событий $E_k^{t+1}, \ k=1,\dots,{\lambda}/2$ при
фиксированной~$\Pi^t$, имеем $p \ge 1-(1-q)^{{\lambda}/2} \ge 1-e^{-q
{\lambda}/2}$, где последнее неравенство следует из леммы~\ref{lemma_lb}.
Оценим снизу вероятность~$q$:
$$
q \ge L \varepsilon \left(1-\left(1-\frac{1}{{\lambda}}\right)^{2s}
\right).
$$
Однако, $(1-1/{\lambda})^{2s} \le (1-1/{\lambda})^{2r{\lambda}} \le 1/{e^{2r}}$ снова по
лемме~\ref{lemma_lb}, поэтому
\begin{equation}\label{bound_on_q}
q \ge L \varepsilon \left(1-\frac{1}{e^{2r}}\right) = Lc,
\end{equation}
где $c:=\varepsilon \left(1-\frac{1}{e^{2r}}\right)$ для краткости. В
дальнейшем мы воспользуемся тем, что из~(\ref{N_condition}) и
(\ref{bound_on_q}) вытекает
\begin{equation} \label{useful}
{\lambda} \ge \frac{2}{L \varepsilon \left(1-1/e^{2r}\right)} \ge 2/q.
\end{equation}
Для оценки снизу вероятности~$p$ применим лемму~\ref{lemma_ub}, из
которой следует что при любом $z \in [0,1]$
\begin{equation}\label{simple} 1-\frac{z}{e} \ge e^{-z}.
\end{equation}
Положим $z=e^{-q {\lambda}/2+1}$. Тогда ввиду неравенства~(\ref{useful}),
$z\le 1$, и следовательно,
\begin{equation}\label{lower_bound_on_p}
p \ge 1-e^{-q {\lambda}/2} \ge \exp\left\{-e^{1-q {\lambda}/2}\right\} \ge
\exp\left\{-e^{1-Lc{\lambda}/2}\right\}.
\end{equation}

От анализа потомков фиксированной популяции~$\Pi^t$ перейдем к
случайной последовательности популяций~$\Pi^0,\Pi^1,\dots$.
Заметим, что $p_{LB}^h$ является оценкой снизу для вероятности
достичь локальный оптимум за серию из не более~$h$ итераций,
улучшающих значение рекорда целевой функции. Действительно,
пусть~$A_{t}=E_1^{t}+\dots+E_{{\lambda}/2}^{t}, \ t=1,2,\dots$. Тогда по
лемме~\ref{lemma_cond},
\begin{equation}\label{cond_probab}
{\bf P}\{A_1\& \dots \& A_h\} = {\bf P}\{A_1\} \prod_{t=1}^{h-1}
{\bf P}\{A_{t+1}|A_1 \& \dots \& A_{t}\} \ge p_{LB}^h.
\end{equation}

Итак, положим $p_{LB}=\exp\left\{-e^{1-Lc{\lambda}/2}\right\}$. Снова
воспользовавшись условием~(\ref{N_condition}), получаем оценку
снизу для вероятности достичь локальный оптимум за серию из не
более~$h$ улучшающих рекорд итераций:
$$
p_{LB}^h = \exp\left\{-h e^{1-Lc{\lambda}/2}\right\} \ge
 \exp\left\{-h e^{-\ln h}\right\}=1/e.
$$
Первая часть утверждения теоремы доказана.

Для оценки среднего времени получения локального оптимума
рассмотрим последовательность серий по~$h$ итераций в каждой.
Пусть событием~$D_i, \ i=1,2,\dots,$ является отсутствие
локального оптимума в популяции GA в $i$-той серии. При выполнении
условий леммы вероятность каждого события~$D_i, \ i=1,2,\dots,$ не
превышает $\mu=1-1/e$ при любой предыстории работы алгоритма. По
аналогии с~(\ref{cond_probab}) заключаем: $ {\bf P}\{D_1\& \dots
\& D_k\} \le \mu^k.$ Таким образом, если через~$Y$ обозначить
случайную величину, равную номеру первой серии, на которой
локальный оптимум будет получен, то, пользуясь свойствами
математического ожидания (см., например,~\cite{Gnedenko}),
получаем
$$ E[Y] = \sum_{i=0}^{\infty} {\bf P}\{Y > i\} =
1+\sum_{i=1}^{\infty} {\bf P}\{D_1\& \dots \& D_i\} \le 1
+\sum_{i=1}^{\infty} \mu^i = e.
$$
Следовательно, локальный оптимум достигается не позднее, чем
за~$eh$ поколений GA в среднем. Q.E.D.\\

Пусть $\lceil \cdot \rceil$ обозначает округление вверх. Тогда в
условиях теоремы, при
\begin{equation}\label{ga_settings}
{\lambda} = 2 \left\lceil \frac{1+ \ln h}{L\varepsilon(1-1/e^{2r})}
\right\rceil, \quad s = \lceil r{\lambda} \rceil,
\end{equation}
обеспечено получение локального оптимума в GA за~$O(h)$ поколений в
среднем.


\begin{exercise} \label{ex:Hamming1} Доказать с использованием леммы~\ref{lemma_ub}, что если расстояние Хэмминга между $n$-битовыми строками ${\bf x}$ и ${\bf y}$ равно~1, то при использовании оператора мутации из КГА, где 
$P_{\rm m}=1/n,$ вероятность того, что $\mbox{Mut}^*({\bf
x})={\bf y},$ оценивается снизу величиной~$1/(e^e n)$.
\end{exercise}

\paragraph{Задача ONEMAX.} В качестве примера применения
теоремы~\ref{th_GA_LS} рассмотрим одноэкстремальную задачу
безусловной оптимизации с целевой функцией $ONEMAX({\bf
x})=\sum_{i=1}^n x_i$ на множестве $\mbox{\rm Sol}=X=\{0,1\}^n$. В
качестве функции приспособленности естественно выбрать $\Phi({\bf
x}) \equiv ONEMAX({\bf x})$. В системе окрестностей, порожденной
метрикой Хэмминга радиуса~1, точка~$(1,1,...,1)$ является
единственным локальным оптимумом, а значит и глобальным.

Пусть в ГА c турнирной селекцией используется операторы мутации и
скрещивания из КГА при $P_{\rm m}=1/n$ и при константной
вероятности $P_{\rm c}<1$. Как следует из упражнения~\ref{ex:Hamming1}, для любого
${\bf x} \in {\rm \mbox{\rm Sol}}$ и любого ${\bf y} \in
\mathcal{N}({\bf x})$ выполнено ${\bf P}\{\mbox{Mut}^*({\bf
x})={\bf y}\} \ge 1/(e^e n)=:L.$ По теореме~\ref{th_GA_LS} заключаем,
что при $s \ge r{\lambda}$, константном $r>0$, и
$$
{\lambda} = \left\lceil \frac{2e^e n(1+ \ln n)}{\varepsilon (1-1/e^{2r})}
\right\rceil,
$$
ГА впервые посещает оптимум в среднем не позднее, чем за~$en$
поколений. Обозначим через $T$ число обращений к функции
приспособленности за время работы ГА до первого получения
оптимального решения. Величину $E[T]$ в англоязычной литературе
принято называть runtime. 
\begin{corollary} \label{cor:onemax}
Пусть в ГА c турнирной селекцией используется операторы мутации и
скрещивания из КГА при $P_{\rm m}=1/n$ и при константной
вероятности $P_{\rm c}<1$, тогда при $s \ge r{\lambda}$, константном $r>0$, и
${\lambda} = \left\lceil \frac{2e^e n(1+ \ln n)}{\varepsilon (1-1/e^{2r})}\right\rceil,$
для функции приспособленности $ONEMAX({\bf x})$ имеем $E[T]=O(n^2\ln n)$.
\end{corollary}
Для сравнения: локальный поиск с окрестностью единичного радиуса Хэмминга получает оптимум в этой задаче
после просмотра $O(n^2)$ пробных решений.

\paragraph{Полиномиально ограниченные задачи комбинаторной оптимизации.}

Задача комбинаторной оптимизации называется {\em полиномиально
ограниченной,} если существует полином от~$|I|$, ограничивающий
значения $f_I({\bf x})$, ${\bf x} \in {\rm \mbox{\rm Sol}}(I)$.
Змаетим, что процедура турнирной селекции требует
времени~$O(s)=O({\lambda})$. Следовательно, имеет место

\begin{theorem}\label{GA_LS}
Если 
\begin{itemize}
\item задача комбинаторной оптимизации~${\Pi=(\mbox{\rm Inst},\mbox{\rm Sol},f_I)}$,
\item трудоемкости операторов $\mbox{Mut}$ и
$\mbox{Cross}$,
\item а также функция $1/L(I)$
\end{itemize}
полиномиально ограничены, то в случае $\mbox{\rm Sol}=\{0,1\}^{n(I)}$, при соответствующем
выборе параметров ГА, локальный оптимум впервые достигается в
среднем за полиномиально ограниченное время.
\end{theorem}

Если семейство окрестностей $\mathcal{N}({\xi})$ порождено
метрикой Хэмминга с константным радиусом окрестности, то
существует оператор мутации~$\mbox{Mut}({\xi})$, вычислимый за
полиномиально ограниченное время и осуществляющий равновероятный
выбор особей-потомков из множества~$\mathcal{N}({\xi})$ при
заданном~${\xi}$. Тогда ${1/L}$ также ограничена сверху некоторым
полиномом от~$|I|$. Таким образом, теорема~\ref{GA_LS} применима
ко многим известным системам окрестностей для задач комбинаторной
оптимизации.

В настоящем разделе не учитывался тот факт, что в результате
действия кроссинговера приспособленность потомков может оказаться
выше приспособленности родителей. Улучшение известных
теоретических оценок для ГА за счет такой возможности
является более сложной задачей, которую пока удалось решить только для
нескольких небольших классов задач дискретной оптимизации.

\chapter{Эволюционные алгоритмы}

\section{Общий вид опраторов ЭА~\cite{ErRe02}} \label{sec:ea}

В дальнейшем нам потребуется символ, обозначающий множество всех
генотипов популяции $\Pi$, т.е. {\em генофонд}. Запишем его как
$\widehat{\Pi}=\cup_{i=1}^{\lambda} \{\xi^i\}$.

В общем случае работа ЭА может быть описана с помощью
операторов, представляющих собой следующие
рандомизированные процедуры, т.е. программы для
вероятностной машины Тьюринга -- см., например, \cite{KSV},
гл.~3.

1. Функция $Terminate$ возвращает <<ложь>>, пока следует
продолжать работу, и <<истина>>,
когда необходимо остановить выполнение ЭА и выдать ответ.\\

2. Оператором селекции $Select: B^{\lambda} \to B^{{\lambda}'}$ извлекается
${\lambda}'$ копий генотипов родителей из текущей популяции,
которые помещаются в промежуточную популяцию
$\Pi'=Select\left(\Pi^t\right)$, причем
$\widehat{\Pi'} \subseteq \widehat{\Pi^t}$.\\

3. Действием оператора воспроизведения $Reproduce: B^{{\lambda}'} \to
B^{{\lambda}''}$ вносятся некоторые случайные изменения в генотипы,
полученные от родительских особей. Таким образом создаются ${\lambda}''$
генотипов-потомков, составляющих популяцию $\Pi''$. (В
частности, в случае КГА ${\lambda}''= {\lambda}'= {\lambda}$, и действие данного
оператора состоит в последовательном применении скрещивания и
мутации.)\\

4. С помощью оператора выживания $Survive: B^{\lambda} \times
B^{{\lambda}''} \to B^{\lambda}$ определяются генотипы из популяции $\Pi^t$
и их потомки из $\Pi''$, которые будут добавляться в
очередную популяцию $\Pi^{t+1}$, т.е.
$\widehat{Survive}\left(\Pi^t,\Pi''\right)
\subseteq \widehat{\Pi^t} \cup \widehat{\Pi''}$.\\

5. Начальная популяция $\Pi^0=Init$ строится случайным образом с
помощью рандомизированной процедуры~$Init$.\\

Работа ЭА начинается со случайной начальной популяции
$\Pi^0=Init$ и продолжается итерациями случайного
отображения
$$\Pi^{t+1}=
Survive
\left(\Pi^t,Reproduce\left(Select\left(\Pi^t\right)\right)\right),
$$
пока не будет выполнено условие остановки $Terminate=$<<истина>>.
Работа заканчивается выводом в качестве ответа лучшего найденного
решения $x(\tilde{\xi}^t)$, где
$$
\tilde{\xi}^t= \mbox{arg max }\{f(x(\xi^{i,\tau})):
\tau=0,...t,i=1,...,{\lambda}\}.
$$

Условие остановки $Terminate$ может быть простым
ограничением по общему числу итераций, либо по числу
итераций без улучшения рекорда целевой функции
$f(x(\tilde{\xi}^t))$. В некоторых задачах можно заранее
определить требуемое значение целевой функции, по
достижению которого алгоритм останавливается. Как правило,
в дальнейшем при теоретическом исследовании алгоритмов для
удобства будем полагать, что условие остановки никогда не
выполняется \footnote{Однако практика показывает, что
многократный независимый перезапуск алгоритма зачастую
позволяет значительно улучшить результаты при том же общем
времени вычислений}. Очевидно, всякий представитель класса
эволюционных алгоритмов может быть реализован на
вероятностной машине Тьюринга.

Распределение вероятностей на выходе процедур $Select$,
$Reproduce$, $Survive$ должно полностью определяться входными
данными решаемой задачи (которую будем считать фиксированной),
номером итерации $t$ и одной или двумя популяциями-аргументами,
поданными на вход процедуры. Таким образом, имеют место {\it
марковские свойства} указанных операторов, которые могут быть формализованы
следующим образом.

Пусть ${\bf M}$ -- вероятностная машина Тьюринга, реализующая
данный ЭА, и $\Theta$ обозначает последовательность состояний,
которые проходит машина~${\bf M}$ до применения рассматриваемого
оператора $Select$, $Reproduce$ или $Survive$.

Условимся обозначать детерминированные популяции или реализации
случайных популяций буквой~$\pi$, а популяции, являющиеся
случайными величинами -- как и прежде, заглавными буквами~$\Pi$
(например, для реализации~$\Pi^t$ будем использовать~$\pi^t$).
Аналогично поступим с другими случайными величинами и их
реализациями ($\xi, \Xi; \ \Theta, \theta$ и т.д.). Тогда:

1) Для любых $\pi' \in B^{{\lambda}'}$, $\pi\in B^{\lambda}$, $t\ge 0$ и любой
последовательности состояний~$\theta$ выполнено равенство
\begin{equation}\label{SelMark}
P\left\{\pi'=Select(\pi)\right\} = P\left\{\pi'=Select(\pi)|
 \Theta=\theta\right\}.
\end{equation}

2) Для любых $\pi''\in B^{{\lambda}''}$, $\pi'\in B^{{\lambda}'}$, $\pi \in B^{\lambda}$,
$t\ge 0$ и последовательности $\theta$, на шаге~$t$ выполнено
равенство
\begin{equation}\label{RepMark}
P\left\{\pi''=Reproduce(\pi')\right\}=
\end{equation}
$$
P\left\{\pi''=
  Reproduce(\pi')|\ \Pi^t=\pi \ \& \ \Theta=
  \theta\right\}.
$$

3) Для любых $\pi \in B^{\lambda}$, $\pi''\in B^{{\lambda}''}$, $\pi'\in B^{{\lambda}'}$,
$\pi^{t+1} \in B^{\lambda}$, $t\ge 0$ и последовательности $\theta$ на
шаге~$t$, выполнено равенство
\begin{equation}\label{SurMark}
P\left\{\pi^{t+1}=Survive(\pi, \pi'')\right\}=
\end{equation}
$$
P\left\{\pi^{t+1}=Survive(\pi, \pi'') | \Pi^t=\pi \ \& \
\Pi''=\pi''\ \& \ \Pi'=\pi'\ \& \ \Theta=\theta\right\}.
$$

Пример: КГА и все рассмотренные ранее его модификации
соответствуют приведенной общей схеме ЭА.

\section{Эволюционные стратегии ($\mu,\lambda$)-ES, ($\mu+\lambda$)-ES}

Один из первых вариантов эволюционной стратегии (1+1)-ES был
предложен Л.А. Растригиным \cite{Ras}, гл.~2 (где этот алгоритм
был назван {\it локальным поиском с пересчетом при неудачном
шаге}). И.~Реченберг  \cite{Re73} сформулировал более общие
вычислительные схемы эволюционных стратегий, которые и приводятся
ниже.

Прежде, чем дать описание алгоритмов, определим один
вспомогательный оператор. 
Пусть оператор~$s_{\mu}$ из данной на
вход популяции генотипов (численностью не менее $\mu$) выбирает
без повторений $\mu$ особей с наибольшей приспособленностью и
возвращает
популяцию из них в качестве результата. Заметим, что если численность 
входной популяции равна $\lambda$, то данный оператор реализует уже рассмотренную
ранее $(\mu,\lambda)$-селекцию.\\

{\bf Общая схема алгоритмов $(\mu,\lambda)$-ES и $(\mu+\lambda)$-ES}\\
1. Построить $\Pi^{(0)}:=(\xi^{1,0},\dots,\xi^{\mu,0}).$\\
2. Для $t:=0$ до $t_{max}-1$ выполнять: \\
\mbox{\hspace{5mm}} 2.1 Для $i:=1$ до $\lambda$ выполнять 2.1.1, 2.1.2:\\
\mbox{\hspace{10mm}}2.1.1 Выбрать $u_i$
с равномерным распределением из $\{1,2,\dots,\mu\}$.\\
\mbox{\hspace{10mm}}2.1.2 Положить $\eta^i:=Mut(\xi^{u_i,t}).$\\
\mbox{\hspace{5mm}} 2.2 Положить $\Pi^{t+1}:= \left \{
   \begin{array}{l}
        s_{\mu}(\eta^1,\dots,\eta^{\lambda}) \mbox{ в алгоритме } (\mu,\lambda)\mbox{-ES}\\
        s_{\mu}(\xi^{1,t},\dots,\xi^{\mu,t},\eta^1,\dots,\eta^{\lambda})
         \mbox{  в алгоритме } (\mu+\lambda)\mbox{-ES}.
   \end{array} \right.$\\
\mbox{\hspace{5mm}} 2.3 $t:=t+1.$\\
3. Результат -- наиболее приспособленный из найденных
генотипов~$\tilde{\xi}^t$.\\


Эволюционные стратегии изначально были предложены для задач
непрерывной оптимизации, где $X\subseteq {\bf R}^n$ имеет мощность
континуума. В таких задачах часто используются операторы мутации
$Mut_{\sigma}$ с нормально распределенным случайным шагом:
$$
Mut_{\sigma}(\xi)=x^{-1}( x(\xi)+Z),
$$
где $Z=(Z_1,...,Z_n)$ и $Z_i$, $i=1,...,n$ -- независимые
нормально распределенные случайные величины со стандартным
отклонением~$\sigma$ (это настраиваемый параметр алгоритма)
и нулевым математическим ожиданием.

При решении задач дискретной оптимизации, как правило, используется оператор
мутации из КГА. В таком случае алгоритмы обозначаются как ($\mu,\lambda$)-EA, ($\mu+\lambda$)-EA,
соответственно.

\section{Сходимость эволюционных алгоритмов}

Введем обозначение для целевой функции от фенотипа лучшей особи на
поколении $t$:
$$
F_t=\max\{f(x(\Xi^{1,t})), f(x(\Xi^{2,t})),...,f(x(\Xi^{{\lambda},t}))\}.
$$

\begin{definition} Будем говорить, что популяция ЭА сходится к
оптимуму в задаче~(1) почти наверное, если $F_t \to f^*$ почти наверное (п.н.) при~${t \to \infty}$.\\
\end{definition}

Для задачи~(2) сходимость популяции ЭА к оптимуму п.н.
определяется аналогично.\\

Поведение ЭА, отвечающего приведенной выше общей схеме, может быть
описано цепью Маркова. Дальнейший анализ ЭА мог бы быть
осуществлен с помощью классической теории конечных марковских
цепей (см., например,~\cite{Rud}). Тем не менее, многие важные
результаты гораздо компактнее могут быть получены непосредственным
изучением ЭА элементарными средствами теории вероятностей.\\

\begin{definition} (классификация операторов ЭА)

1. Оператор селекции $Select$ будем называть {\it невырожденным, }
если существует такое $\epsilon_1 >0$, что при любых $\pi^t\in
B^{\lambda}, \xi \in \widehat{\pi^t}$, $t\ge 0$
$$
P\{\xi \in \widehat{Select}(\pi^t)\}\ge \epsilon_1.
$$

2. Пусть $B^*$ обозначает множество оптимальных генотипов.
Оператор воспроизведения $Reproduce$ будем называть {\it
связывающим}, если существует $\epsilon_2 >0$, такое что для любых
$\xi\in B$, $t \ge 0$ найдется последовательность генотипов
$\eta^0, \eta^1,...,\eta^{k(\xi)}$, где $\eta^0=\xi$,
$\eta^{k(\xi)}\in B^*$ и при всех $i=0,...,k-1$ имеем
$$
P\{\eta^{i+1} \in \widehat{Reproduce}(\pi')\}\ge \epsilon_2 \ \
\forall \ \pi': \eta^i \in \widehat{\pi'}.
$$

3. Оператор выживания $Survive$ будем называть {\it
невырожденным}, если существует $\epsilon_3 >0$, такое что для
любого $t\ge 0$
$$
P\{\xi \in \widehat{Survive}(\pi^t,\pi'')\}\ge \epsilon_3 \ \
\forall \ \pi^t\in B^{\lambda}, \pi''\in B^{{\lambda}''}: \xi \in \widehat{\pi''}.
$$

4. Оператор выживания $Survive$ будем называть {\it
консервативным,} если для любых $\pi^t,\pi''$, $t\ge 0$
выполняется
$$
\max\{f(x(\xi)):\xi\in \widehat{Survive}(\pi^t,\pi'')\} \ge
\max\{f(x(\xi)):\xi\in \widehat{\pi^t} \cup \widehat{\pi''}\}.
$$

\end{definition}

В данном определении, виду марковских свойств, мы пользовались
обычными вероятностями для результатов случайных операторов,
вместо условных вероятностей, зависящих от предыстории работы
алгоритма.

\begin{prop} \label{condp}
(об условных вероятностях) Для любых событий $A,A',A''$
$$
P\{A\ \& \ A'|A''\}=P\{A|A'\ \& \ A''\} P\{A'|A''\},
$$
если эти условные вероятности определены.
\end{prop}
Доказывается трехкратным применением формулы из определения условной вероятности.\\

\begin{prop} \label{totcond}
(формула полной условной вероятности) Для любых событий $A,A'$ и
альтернатив $A_1,...,A_k$ таких, что $A_i \cap A_j=\emptyset, i
\neq j$ и $A'=\cup_{i=1}^k A_i$, выполнено равенство
$$
P\{A|A'\}=\sum\limits_{i=1}^k P\{A|A_i\} P\{A_i|A'\},
$$
если эти условные вероятности определены.
\end{prop}
{\bf Доказательство.}
$$
P\{A|A'\}=\frac{P\{A \ \& \ A'\}}{P\{A'\}}= \sum\limits_{i=1}^k
\frac{P\{A \ \& \ A' \ \& \ A_i\}}{P\{A'\}}= \sum\limits_{i=1}^k
\frac{P\{A| A' \ \& \ A_i\} P\{A' \ \& \ A_i\}}{P\{A'\}}.
$$
Пользуясь определением условной вероятности и тем, что $A' \ \& \
A_i = A_i$ для всех $i=1,..,k$, приходим к требуемому равенству.
Q.E.D.

Следующая теорема о непрерывности вероятностной меры известна из
курса теории вероятностей.

\begin{theorem} \label{cont_mesur}
Пусть $\{A_n\}$ -- последовательность множеств из сигма-алгебры
событий, $A_{n+1} \subseteq A_n \ \forall n$, тогда
$$
\lim_{n \to \infty} P(A_n)= P(\cap_{n=1}^{\infty} A_n).
$$
\end{theorem}

При кодировке решений задачи (\ref{discrprb}) оптимум может не
быть представлен в пространстве генотипов. В следующей теореме
такая ситуация исключается предположением, что $B^*\neq
\emptyset$.

\begin{theorem} \label{convea}
о сходимости ЭА (Айбен, Аартс, ван Хи, 1989) \cite{EAH91,Rud}.
Пусть $B^*\neq \emptyset$ и функция
$Terminate$ никогда не возвращает значение~<<истина>>. Тогда\\

1. В случае невырожденной селекции и выживания, при связывающем
операторе воспроизведения имеем
\begin{equation}\label{finconv}
P\{\exists t : \widehat{\Pi^t} \cap B^* \neq \emptyset\} =1
\end{equation}
(т.е. в ЭА оптимальный генотип порождается с вероятностью единица
за конечное число итераций).

2. В случае выполнения равенства (\ref{finconv}) при
консервативном операторе выживания популяция ЭА сходится к
оптимуму почти наверное.
\end{theorem}

{\bf Доказательство.}

1. Рассмотрим популяцию $\Pi^0$ как случайную величину. Как будет
видно в дальнейшем, вместо $\Pi^0$ можно было бы взять популяцию
$\Pi^t$ на любой другой итерации -- это только усложнило бы
обозначения. Пусть $\pi^0$ -- некоторая вспомогательная (не
случайная) популяция и $\xi^{1}$ -- первый ее генотип.

Пусть $k(\xi^{1})$-- число элементов в пути $\eta^0=\xi^{1},
\eta^1,...,\eta^{k(\xi^{1})}$ от генотипа $\xi^{1}$ до $B^*$ из
определения связывающего оператора воспроизведения, и пусть
$k^*=\max\{ k(\xi): \xi \in B\}$. Очевидно, $k^*$ конечно, т.к.
$B$ конечно.

Обозначим через $U(\xi^{1},i)$ событие $\{\eta^{0} \in
\widehat{\Pi^{0}}, \eta^{1} \in \widehat{\Pi^{1}},...,\eta^{i} \in
\widehat{\Pi^{i}}\}$. Рассмотрим условную вероятность
$p(i)=P\{\eta^{i+1} \in \widehat{\Pi^{i+1}} | U(\xi^{1},i) \ \& \
\Pi^0=\pi^0\}$ при наличии генотипов $\eta^{j} \in
\widehat{\Pi^{j}}$ на всех итерациях $j$ от 0 до $i-$той
включительно, получить в новой популяции $\Pi^{i+1}$ следующий
генотип $\eta^{i+1}$. Здесь $i$ принимает значения от~0 до
$k(\xi^{1})-1$. Покажем, что $p(i)>0.$

С учетом Утверждения \ref{condp} об условных вероятностях имеем:

$$
p(i) \geq P\{\eta^{i+1} \in \widehat{\Pi^{i+1}} \ \& \ \eta^{i+1}
\in \widehat{\Pi''} \ \& \  \eta^i \in \widehat{\Pi'}
|U(\xi^{1},i) \ \& \ \Pi^0=\pi^0\}=
$$
\begin{equation} \label{long}
=p_{\rm surv} \cdot  p_{\rm sel \ \& \ rep},
\end{equation}
где
$$
p_{\rm surv}=P\{\eta^{i+1} \in \widehat{\Pi^{i+1}} | \eta^{i+1} \in
\widehat{\Pi''} \ \& \ \eta^i \in \widehat{\Pi'} \ \& \
U(\xi^{1},i) \ \& \ \Pi^0=\pi^0\},
$$
$$
p_{\rm sel \ \& \ rep}= P\{\eta^{i+1} \in \widehat{\Pi''} \ \& \  \eta^i \in
\widehat{\Pi'} | U(\xi^{1},i) \ \& \ \Pi^0=\pi^0\}.
$$

По Утверждению \ref{condp},
$$
p_{\rm sel \ \& \ rep}= P\{\eta^{i+1} \in \widehat{\Pi''} | \eta^i \in \widehat{\Pi'}
\ \& \ U(\xi^{1},i) \ \& \ \Pi^0=\pi^0\} \cdot
$$
\begin{equation}\label{eqn:p_rep}
\cdot P\{\eta^i \in
\widehat{\Pi'}| U(\xi^{1},i) \ \& \ \Pi^0=\pi^0\}.
\end{equation}

Покажем, что последний сомножитель положителен. Заметим, что 
по утверждению \ref{totcond}
$$P\{\eta^i \in
\widehat{\Pi'}| U(\xi^{1},i) \ \& \ \Pi^0=\pi^0\}=
$$
$$
=\sum\limits_{\pi: \ \eta^i \ \in \widehat{\pi}} P\{\eta^i \in
\widehat{Select(\Pi^i)} | \Pi^i=\pi \} P\{\Pi^i=\pi| U(\xi^{1},i)\
\& \ \Pi^0=\pi^0\}\geq
$$
$$
\ge \epsilon_1 \cdot \sum\limits_{\pi \ : \ \eta^i \in \widehat{\pi}}
P\{\Pi^i=\pi| U(\xi^{1},i) \ \& \ \Pi^0=\pi^0\}=
$$
$$
=\epsilon_1 \cdot \sum\limits_{\pi \ : \ \eta^i \in \widehat{\pi}}
\frac{P\{\Pi^i=\pi \ \& \ U(\xi^{1},i) \ \& \ \Pi^0=\pi^0\}}
{P\{U(\xi^{1},i) \ \& \ \Pi^0=\pi^0\}}=\epsilon_1.
$$
Здесь суммирование ведется только по тем $\pi$, которые содержат
$\eta^i$ ввиду того, что оператор селекции может построить
популяцию с $\eta^i$ только если особь~$\eta^i$ имелась во входной
популяции. Итак, положительность последнего сомножителя в~(\ref{eqn:p_rep}) доказана. 

Аналогичным способом, используя утверждение~\ref{totcond} и
определение связывающего воспроизведения, получаем ${
P\{\eta^{i+1} \in \widehat{\Pi''} | \eta^i \in \widehat{\Pi'} \ \&
\ U(\xi^{1},i) \ \& \ \Pi^0=\pi^0\} }\ge\epsilon_2$. Следовательно,
$p_{\rm sel \ \& \ rep} \ge \epsilon_2 \epsilon_1$.

Из определения невырожденного оператора выживания с использованием
утверждения~\ref{totcond} также вытекает, что $p_{\rm surv} \ge
\epsilon_3$. Таким образом, существует достаточно малая
положительная величина $\delta=\epsilon_1 \epsilon_2 \epsilon_3$,
такая что $0<\delta < p(i)$.


Далее, пусть $p^*(\pi^0)$ будет условной вероятностью при
$\Pi^0=\pi^0$ попасть в $B^*$ на  итерации~$k(\xi^{1})$, следуя за
цепочкой генотипов из определения связывающего воспроизведения:
$\eta^0=\xi^{1}, \eta^1 \in \widehat{\Pi^1}, \eta^2 \in
\widehat{\Pi^2},..., \eta^{k(\xi^{1})} \in
\widehat{\Pi^{k(\xi^{1})}} \cap B^*$.

Заметим, что по лемме~\ref{lemma_cond} для любых событий
$A_0,A_1,...,A_n$,
$$
P\{A_0\& A_1\& ... \& A_n\} = P\{A_0\} \prod_{i=0}^{n-1}
P\{A_{i+1}|A_0 \& A_1\& ... \& A_i\}.
$$

Таким образом,

\begin{equation}\label{iter}
p^*(\pi^0)=\prod\limits_{i=0}^{k(\xi^{1})-1} P\{\eta^{i+1} \in
\widehat{\Pi^{i+1}} |U(\xi^{1},i)
 \ \& \ \Pi^0=\pi^0\}.
\end{equation}

Оценивая $p^*(\pi^0)$ с помощью $\delta$, имеем: $p^*(\pi^0)\geq
\delta^{k(\xi^{1})}.$ Пусть $\Delta=\delta^{k^*}$. Тогда
$\min_{\pi^0 \in B^{\lambda}} p^*(\pi^0)\geq \Delta$, и
\begin{equation}\label{Less1}
\max_{\pi^0 \in B^{\lambda}} P\{F_t<f^*, t=0,...,k(\xi^1)
|\Pi^0=\pi^0\}\leq \max_{\pi^0 \in B^{\lambda}} (1-p^*(\pi^0))\leq
1-\Delta<1.
\end{equation}

Обозначим через $A(\vartheta)$ событие, состоящее в отсутствии
оптимальных генотипов до итерации~$\vartheta$ включительно. С
учетом леммы~\ref{lemma_cond}, марковости операторов ЭА и формулы~(\ref{Less1}), для
любого $s$ имеем \linebreak
%
%
$P\{A(s\cdot(k^*+1))\}\leq P\{A((s-1)(k^*+1))\}(1-\Delta)$, откуда
по индукции заключаем, что
$$
P\{A(\vartheta)\} \le (1-\Delta)^{\lfloor \vartheta / (k^*+1)
\rfloor}.
$$
Тогда с использованием теоремы~\ref{cont_mesur} о непрерывности
вероятностной меры получаем:
$$
P\{\widehat{\Pi^t} \cap B^* = \emptyset \ \forall t\}=
P\left\{\bigcap\limits_{\vartheta=0}^{\infty}
A(\vartheta)\right\}=
$$
$$
= \lim\limits_{\vartheta \to \infty} P\left\{A(\vartheta)\right\}
\le \lim\limits_{\vartheta \to \infty} (1-\Delta)^{\lfloor
\vartheta /(k^*+1) \rfloor} =0.
$$
2. В случае, если для любых $\pi,\pi''$
$$
\max\{f(x(\xi)):\xi\in \widehat{Survive}\left(\pi,\pi''\right)\}
\ge \max\{f(x(\xi)):\xi\in \widehat{\pi} \cup \widehat{\pi''}\},
$$
после обнаружения оптимального генотипа, в каждой популяции
$\Pi^t$ будет присутствовать такой генотип, и значит, порождение
оптимума за конечное число итераций обеспечивает сходимость
популяции ЭА к оптимуму п.н. Q.E.D.

\begin{definition}
Оператор воспроизведения $Reproduce(\Pi')$ будем
называть {\it положительным}, если существует такое $\epsilon>0$,
что для любых $\Pi'\in B^{{\lambda}'}$, $\eta \in B$ и $t\ge 0$
$$
P\{\eta \in \widehat{Reproduce(\Pi')}\} \ge \epsilon.
$$
\end{definition}

\begin{note} \label{note_EAs} Пусть $B^* \ne \emptyset$ и функция $Terminate$
никогда не возвращает значение <<истина>>. Тогда положительности
воспроизведения и консервативности выживания достаточно для
сходимости популяции ЭА к оптимуму почти наверное.
\end{note}

{\bf Доказательство} аналогично п.1 теоремы \ref{convea}, т.к.,
полагая $k(\xi^{1})=1$, ввиду положительности воспроизведения
имеем
$\delta>0$ для любой $\pi^0 \in B^{{\lambda}}$.\\

{\bf Примеры.}  Предположение  $B^* \neq \emptyset$ будем считать выполненным.\\

1. КГА при $0<p_m<1$ обладает положительным оператором
воспроизведения, т.к. при мутации из любого генотипа с ненулевой
вероятностью может быть
получен любой генотип.\\

2. Если в КГА $p_m=0$ или $p_m=1$, то оператор воспроизведения КГА
не является связывающим, и можно привести примеры начальных
популяций, начиная с
которых КГА не сможет никогда найти некоторые решения.\\

3. Если в КГА оператор мутации заменить на одноточечную мутацию в
случайно выбранном гене (с равномерным распределением), то
оператор воспроизведения будет связывающим и будет применим п.1
теоремы о
сходимости ЭА.\\

4. К КГА неприменим п.2 теоремы о сходимости ЭА, т.к. оператор
выживания не является консервативным и однажды найденное
оптимальное решение может быть потеряно. Однако, это свойство
может быть достаточно легко обеспечено,
например, как это делается в ГА с элитой.\\

5. Популяция $(\mu+\lambda)$-ES или $(\mu+\lambda)$-EA с оператором мутации, который выдает 
генотип из~$B^*$ с ненулевой вероятностью (например, при мутации с
нормально распределенным шагом), сходится к оптимуму почти
наверное, ввиду замечания~\ref{note_EAs}.\\


{\bf Замечание.} Свойства сходимости, полученные в последней
теореме (хотя бы в смысле п.1, как в КГА), являются желательными
для всякого ЭА. Однако даже наличие такой сходимости не дает
гарантии "надежной работы"\ алгоритма, т.к. на практике число
выполняемых итераций за реальное время может оказаться слишком
мало, чтобы вероятность получения оптимума стала близка к 1.

Как видно из примера~5, даже самый примитивный алгоритм (1+1)-EA с
оператором мутации, имеющим равномерное распределение на множестве
генотипов, обладает всеми свойствами сходимости в смысле теоремы
\ref{convea}. Наконец, детерминированный полный перебор генотипов
всегда за конечное время находит лучший генотип, а значит,
является сходящимся методом (как детерминированный алгоритм).

Из последнего замечания вытекает необходимость более детального
исследования скорости сходимости ЭА к оптимуму или времени первого достижения оптимума 
с учетом других свойств задачи, например, как в следствии~\ref{cor:onemax}. 

\section{Алгоритмы генетического программирования}

Алгоритмы генетического программирования~(ГП) были предложены
Н.Л.Крамером~\cite{Cramer85} и развиты далее в работах Дж.
Козы~\cite{Koza} и других авторов. Идея ГП заключается в том, что
в отличие от ГА, здесь все операции производятся не над строками,
а над деревьями. При этом используются операторы, аналогичные
селекции, скрещиванию и мутации ГА. С помощью деревьев
предлагается кодировать программы для ЭВМ и математические формулы
- таким образом можно организовать эволюцию программного кода для
решения программистской задачи или поиск подходящей функции в
аналитическом виде.

Можно считать, что в ГП фенотипом является программа,
представленная как дерево с терминальными (листья дерева) и
функциональными элементами (все прочие вершины). Терминальные
элементы соответствуют константам, действиям и функциям без
аргументов, а функциональные - функциям, использующим аргументы.

Например рассмотрим функцию x*3/5-1. Терминальные элементы здесь:
$\{x,3,5,1\}$, функциональные: $\{+,*,/\}$.


Схема ГП аналогична схеме ГА, однако операторы скрещивания и
мутации имеют отличия.

В операторе скрещивания выбираются случайные поддеревья
родительских генотипов и происходит обмен.

При построении начальной популяции и в операторах воспроизведения уделяется внимание глубине, ширине и структуре формируемых деревьев~\cite{PP00,Koza}.
Например для
предотвращения чрезмерного разрастания дерева в высоту вводится ограничение
на максимальное количество функциональных элементов в дереве или
максимальную его глубину. Если при скрещивании двух деревьев один
из потомков не удовлетворяет такому ограничению, вместо него
копируется родительское дерево.

При действии оператора мутации случайно удаляется часть дерева и
вместо нее генерируется новое поддерево случайным образом. В
некоторых случаях мутация сводится к случайному изменению
терминальных элементов (тогда для каждого типа элементов должно
быть задано распределение вероятностей, определяющее случайные
изменения).

ГП может рассматриваться как частный случай ГА с изменяющейся
длиной кодировки и специфическими операторами кроссинговера и
мутации, поэтому для них можно применять теорему о сходимости и
доказывать аналоги теоремы о схемах.

Алгоритмы ГП используются для решения задач построения нелинейных моделей (математических выражений, функций, алгоритмов, программ) на основе заданных экспериментальных данных, множества переменных, базовых функций и операций (например для предсказания уровня воды в водохранилище). Также алгоритмы ГП применяются в синтезе решающих деревьев и в некоторых других методах машинного обучения.

\subsection{Кроссинговер в генетическом программировании}

Работоспособность алгоритма ГП существенно зависит от выбора оператора кроссинговера, где родительские решения (деревья) обмениваются своими признаками. Наиболее распространенными вариантами оператора кроссинговера являются одноточечный кроссинговер~\cite{Dhaeseleer} и равномерный кроссинговер~\cite{PP00}.
При одноточечном кроссинговере в родительских решениях выбираются узлы и осуществляется обмен поддеревьями с корневыми вершинами в выбранных узлах. 
Равномерный кроссинговер характеризуется тем, что значение каждого из узлов потомка наследуется из соответствующих значений в родительских деревьях с заданной вероятностью.

Известны различные вариации указанных операторов кроссинговера, где учитываются координаты и значения вершин поддеревьев, их размеры и другие свойства (см., например,~\cite{L2000,MKJ12}).

\section{Алгоритм поиска с запретами (tabu search)}

Алгоритм поиска с запретами предложен Ф. Гловером (F.Glover) в
1986г. (см.~\cite{Kochet,Pirl}). В классическом алгоритме
локального поиска всякий раз выбирается лучшая точка в
окрестности, или точка, имеющая значение целевой функции не хуже,
чем текущая. Если текущая точка является локальным оптимумом, то
на следующей итерации, вероятен возврат в этот же локальный
оптимум. В алгоритме поиска с запретами разработан механизм
предотвращения таких возвратов, основаный на использовании списка
$\Lambda$ из последних $L$ решений:
$\Lambda=(\ell^1,\dots,\ell^L)=({\bf y}^{t-L},..., {\bf
y}^{t-1})$. Множество запрещенных решений при текущем решении
${\bf y}^t$ и списке $\Lambda$ будем обозначать через $Tabu({\bf
y}^t,\Lambda)$.

Как показывает практика, часто удобнее хранить список запрещенных
"ходов"\ (tabu list), т.е. список некоторых способов изменения
 решений, которые приводят к запрещенным решениям.

Примеры запрещенных "ходов": недавно использованное увеличение
(или уменьшение) некоторой координаты; переход в решение со
значением целевой функции, присутствующим в списке
$f(\ell^1),..., f(\ell^L)$; то и другое одновременно и т.д.

Было бы странно, если список запретов не позволял переходить в
заведомо лучшие решения, по сравнению с найденными ранее. Для
того, чтобы этого не происходило, используется "условие
стремления"\ (aspiration condition): если целевая функция в
некоторой точке~${\bf x}$ из текущей окрестности выше значения
целевой функции лучшего из найденных прежде решений, то следует
разрешить переход в решение ${\bf x}$, независимо от $\Lambda$.
Множество решений, попадающих в условия "стремления"\ при текущем
рекорде $\tilde{f}$ будем обозначать через $Asp(\tilde{f})$.

Часто имеется некоторый способ выделения подокрестности ${\cal
N}'({\bf y}^t)\subseteq {\cal N}({\bf y}^t)$ в окрестности ${\cal
N}({\bf y}^t)$ (детерминированно или случайно). Это позволяет
снизить
трудоемкость поиска и "разнообразить"\ его. \\

{\bf Общая схема алгоритма поиска с запретами}\\
1. Выбрать ${\bf y}^1$ случайным образом.\\
2. Положить $\tilde{{\bf y}}:={\bf y}^1$; $\tilde{f}:=f({\bf y}^1)$.\\
3. Инициализировать список $\Lambda=(\ell^1,\dots,\ell^L):=({\bf y}^1,{\bf y}^1,...,{\bf y}^1)$;\\

Итерация $t$.\\
4. Положить $f':=-\infty; {\bf y}':= {\bf y}^t$.\\
5. Для всех ${\bf y}\in {\cal N}'({\bf y}^t) \backslash (Tabu({\bf
y}^t, \Lambda) \backslash Asp(\tilde{f}))$:

\hspace{5em} если $f({\bf y})>f'$, то положить$f':=f({\bf y}), {\bf y}':={\bf y}$.\\
6. Положить ${\bf y}^{t+1}:={\bf y}'$.\\
7. Если $f'>\tilde{f}$, то положить $\tilde{f}:=f({\bf y}'), \tilde{{\bf y}}:={\bf y}'$.\\
8. Обновить список $\Lambda$: исключить первый элемент, сдвинуть
все
элементы влево на одну позицию, положить $\ell^L:={\bf y}^t$.\\
9. Положить $t:=t+1$.\\
10. Если $Terminate=$<<ложь>>, то перейти на шаг 4.\\

{\bf Замечание.} Алгоритм поиска с запретами -- частный случай ЭА,
где ${\lambda}=L+2$ и оператор выживания является консервативным.\\


{\bf Вероятностный поиск с запретами (PTS).}

Вероятностный поиск с запретами предложен Гончаровым Е.~Н. и
Кочетовым Ю.~А.~\cite{GonKoch2002}. Пусть $D=X=\{0,1\}^n$ --
множество допустимых решений. ${\cal N}({\bf x})=\{{\bf y}\in D :
\rho({\bf x},{\bf y}) \leq d\}$, где $\rho({\bf x},{\bf y})$ -
расстояние в метрике Хэмминга. Пусть $P\{{\bf x}' \in {\cal
N}'({\bf x})\}=p$ -- заданная константа для любых ${\bf x}\in D,
{\bf x}'\in {\cal N}({\bf x})$. (Очевидно, в таком случае
$P\{{\cal N}'({\bf x})=\emptyset\}=(1-p)^{|{\cal N}({\bf x})|}>0$,
если $p<1$.)
$$
Tabu({\bf x}, \Lambda)=
$$
$$
\left\{{\bf y}\in D : \ \exists \  k \in \{1,...,L\}: \ |({\bf
y}-{\bf x})_i|=|(\ell^{k+1}-\ell^{k})_i| \ \forall \
i=1,..,n \right\},
$$
где под $\ell^{L+1}$ понимается текущее решение ${\bf x}$.

Настраиваемые параметры: $p,d,L$. Как правило, $d=1$ или 2.

\begin{theorem} О сходимости вероятностного поиска с запретами
\cite{GonKoch1999}. Пусть $p\in (0,1)$. Тогда в вероятностном
поиске с запретами $f({\bf y}^t) \to f^*$ при $t\to \infty$ почти
наверное.
\end{theorem}

Если при пустой подокрестности ${\cal N}'({\bf y}^t)$ список
запретов оставлять без изменений и $0 <L< (n-1)n/4$ , то указанная
сходимость также имеет место~\cite{GonKoch2002}.

\section{Оценки доли особей с заданной приспособленностью}
\label{sec:fractions}

Настоящий раздел посвящен оценкам численности особей достаточно
высокого качества на заданной итерации ЭА с
 полной заменой популяции
и турнирной селекцией, без оператора кроссинговера. Для анализа ЭА
предлагается математическая модель эволюционного процесса,
возникающего при его работе. С использованием данной модели
строятся нижние и верхние оценки среднего числа особей с
приспособленностью не ниже заданного порогового значения.

В литературе предложено несколько подходов к анализу ЭА с
использованием цепей Маркова. Например, при отождествлении
состояний цепи со всевозможными векторами генотипов популяции~$X$
возникает цепь Маркова с $2^{l\lambda}$ состояниями, если ${\mathcal
B}=\{0,1\}^l$. 

В другой модели~\cite{NV}, представляющей работу КГА с помощью цепи
Маркова, все популяции, отличающиеся лишь способом упорядочения
особей, считаются эквивалентными. Каждое состояние можно
представить в виде вектора, имеющего $2^l$ компонент, где каждая
координата отвечает одному из возможных генотипов. Значение
компоненты равно доле, которую составляют в популяции особи с
соответствующим этой компоненте генотипом. В рамках данной модели
найдена переходная матрица цепи Маркова, отвечающей ГА с полной
заменой популяции и пропорциональной селекцией для некоторых
вариантов кроссинговера, и получен ряд важных
результатов, описывающих поведение ГА~\cite{NV,Vo}.

Основные трудности практического применения указанных моделей для
анализа задач комбинаторной оптимизации вызваны необходимостью
использования априорной информации о приспособленности каждого
генотипа из~$\mathcal B$ и большой размерностью матрицы перехода
цепи Маркова. Ниже рассматривается способ
преодоления этих трудностей посредством группировки состояний.
Полученная модель не является цепью Маркова, однако позволяет
найти нижние и верхние оценки числа особей с приспособленностью не
ниже заданного уровня на итерации~$t$.

\subsection{Модель эволюционного процесса}

В настоящем параграфе предлагается модель эволюционного процесса,
возникающего при работе ЭА с полной заменой популяции при
турнирной селекции и отсутствии кроссинговера. Для
применения данной модели не требуется непосредственного
использования информации о приспособленности каждого генотипа, как
в~\cite{Rud,NV}, однако предполагается, что известны некоторые
оценки параметров оператора мутации.

При отсутствии кроссинговера в ЭА с полной заменой
популяции нет необходимости генерировать особи каждой новой
популяции обязательно парами, как в КГА. В
связи с этим для большей общности здесь будем предполагать, что
условие о четности~${\lambda}$ отсутствует и ЭА имеет следующую схему.

{\samepage
\begin{myalgorithm}
{\bf ЭА с полной заменой популяции}
    \label{alg:ga_simple}
\end{myalgorithm}
\noindent {\bf 1.} Положить $t:=0$.\\
{\bf 2.} Для $k$ от 1 до ${\lambda}$ выполнять: \\
{\bf 2.1.} Построить случайным образом особь $\xi^{k,0}$.

 {\bf Итерация~$t$.}\\
\noindent{\bf 3.} Для $k$ от 1 до ${\lambda}$ выполнять шаги 3.1--3.2: \\
{\bf 3.1.} Селекция: выбрать особь $\eta^{kt}:=\xi^{{\rm Sel}(P^t),t}$. \\
{\bf 3.2.} Мутация: положить $\xi^{k,t+1} := \mbox{\rm Mut}(\eta^{kt})$.\\
{\bf 4.} Положить $t:=t+1$.\\
{\bf 5.} Если условие остановки выполнено, то идти на шаг~6,
иначе -- на шаг~3.\\
{\bf 6.} Результатом работы является особь с наибольшей
приспособленностью среди найденных за все итерации. }\\

Пусть $\Phi_0=\min_{\xi \in {\mathcal B}} \Phi(\xi)$ и заданы
линии уровня функции приспособленности $\Phi_1, \ldots , \Phi_d,$
такие что
$$
\Phi_0 < \Phi_1 < \Phi_2 \ldots < \Phi_d.
$$
Число линий уровня и соответствующие им значения функции
приспособленности (кроме~$\Phi_0$) могут быть выбраны достаточно
произвольно с учетом свойств решаемой задачи и оператора мутации.
Далее будем предполагать, что число линий уровня, а также их
значения некоторым образом зависят от индивидуальной задачи~$I$.
Для простоты обозначений линий уровня символ~$I$, как правило, не
будет указываться при их записи.

Сопоставим выбранным линиям уровня следующие подмножества:
$$
H_i=\{\xi : \Phi(\xi) \geq \Phi_i \}, \hspace{1em} i=0,\ldots,d.
$$
Заметим, что $H_0={\mathcal B}$. Для удобства обозначений положим
$H_{d+1}=\emptyset$.

Пусть для всех $i=0,...,d$ и $j=1,...,d$ априори известны нижние
$\alpha_{ij}$ и верхние $\beta_{ij}$ оценки вероятности перехода
из $H_i \backslash H_{i+1}$ в $H_j$ при действии оператора
мутации, а именно для любого $\xi \in H_i \backslash H_{i+1}$
$$
\alpha_{ij} \leq {\bf P}\{{\rm Mut}(\xi)\in H_j\} \leq \beta_{ij},
$$
где
$$
{\bf P}\{{\rm Mut}(\xi)\in H_j\}=\sum\limits_{\xi'\in H_j} {\bf
P}\{{\rm Mut}(\xi)=\xi'\}.
$$
Обозначим матрицу с элементами $\alpha_{ij}$, где $i=0,...,d,
j=1,...,d$ через $\bf A$; аналогичную матрицу верхних оценок
обозначим через $\bf B$.

Представим популяцию на шаге~$t$ при помощи {\it вектора
популяции}
$$
{\bf z}^{(t)}=(z_1^{(t)},z_2^{(t)},\ldots,z_d^{(t)}),
$$
где $z_i^{(t)}\in \R$ - доля генотипов из $H_i$ в популяции
$X^{t}$. Таким образом, ${\bf z}^{(t)}$ -- случайный вектор,
причем $z_i^{(t)}\geq z_{i+1}^{(t)}$ для $i=1,...,d-1$, \since
$H_{i+1} \subseteq H_i$. Далее для  удобства будем полагать
$z_{d+1}=z_{d+1}^{(t)}=0$ при любом~$t$.

Заметим,
что все особи поколения~$t$ в рассматриваемом алгоритме
генерируются с одним и тем же распределением вероятностей.
Следовательно, ${\bf
P}\{\xi^{1t}\in H_j\}=...={\bf P}\{\xi^{\lambda t)}\in H_j\}$ при
любом $t>0$ и $j=1,...,d$. Поэтому далее в выражениях вида ${\bf
P}\{\xi^{kt}\in H_j\}$ номер $k$ не имеет значения и мы полагаем $k=1$. 

\begin{proposition}
\label{vose} При любом $t > 0$ имеет место равенство
${\E[z_i^{(t)}]={\bf P}\{\xi^{(1t)} \in H_i\}}$.
\end{proposition}
{\bf Доказательство.} Рассмотрим последовательность одинаково
распределенных случайных величин ${\mathcal I}_1^i,{\mathcal
I}_2^i,...,{\mathcal I}_{\lambda}^i$, где ${\mathcal I}_k^i=1$, если $k$-я
особь популяции $X^{t}$ лежит в $H_i$, и ${\mathcal I}_k^i=0$ в
противном случае. По определению вектора популяции имеем
$z_i^{(t)}=\sum_{k=1}^{\lambda} {\mathcal I}_k^i/{\lambda}$, следовательно,
$$
\E[z_j^{(t)}]=\sum\limits_{k=1}^{\lambda} \frac{\E[{\mathcal
I}_k^i]}{{\lambda}}= \sum\limits_{k=1}^{\lambda} \frac{{\bf P}\{\xi^{(kt)}
\in H_i\}}{{\lambda}}={\bf P}\{\xi^{(1t)} \in H_i\}. \ \ \Box
$$

Получим нижние и верхние оценки доли особей с генотипами из
подмножеств $H_i$ для $i=1,\ldots,d$ на заданной итерации.

По схеме рассматриваемого ЭА, вероятность того, что выбранный селекцией генотип~$\eta^{kt}$ принадлежит 
какому-либо множеству $S \subset B$, одинакова для всех генотипов популяции на итерации~$t$ и
не зависит от номера~$k$. Поэтому достаточно рассматривать $k=1.$
При векторе текущей популяции~${\bf z}^t=\bf z$ и размере турнира $s$ имеем ${\bf P}\{\eta^{1t} \in H_i \ | \ {\bf z}^t={\bf
z}\}=1-(1-z_i)^s$, и
$${\bf P}\{\eta^{1t} \in H_i\backslash H_{i+1} \ | \ {\bf z}^t={\bf z}\}=$$ 
$$
={\bf P}\{ \eta^{1t} \in  H_i \ | \ {\bf z}^t= {\bf
z}\}-{\bf P} \{\eta^{1t} \in  H_{i+1} \ | {\bf z}^t= \ {\bf z}\}= 
$$
$$
=(1-z_{i+1})^s-(1-z_i)^s.
$$

Если текущая популяция представляется вектором ${\bf z}^{(t)}={\bf z}$, то
условная вероятность того, что очередной генотип,
полученный в результате селекции и мутации, принадлежит
$H_j$, имеет вид:
\begin{equation}
\label{t} {\bf P}\{\xi^{(t+1)} \in H_j |{\bf z}^{(t)}={\bf z}\}=
\sum\limits_{i=0}^d \hspace{0.5em}\sum\limits_{\xi \in
H_i\backslash H_{i+1}} {\bf P}\{{\rm Mut}(\xi) \in H_j \} {\bf
P}\{\eta^{1t} = \xi \ | \ {\bf z}^{(t)} = {\bf z}).
\end{equation}

\subsection{Нижние оценки для случая  турнирной селекции}
Из формулы (\ref{t}) и определения оценок $\alpha_{ij}$ следует,
что
$$
{\bf P}\{\xi^{(t+1)} \in H_j |{\bf z}^{(t)}={\bf z}\}\geq
\sum\limits_{i=0}^d \alpha_{ij} \sum\limits_{\xi \in
H_i\backslash H_{i+1}} {\bf P}\{\eta^{1t} =\xi \ | \ {\bf z}^{(t)}={\bf z}\} =
$$
$$
=\sum\limits_{i=0}^d \alpha_{ij} {\bf P} \{ \eta^{1t} \in H_i\backslash H_{i+1} \ | \ {\bf z}^{(t)}={\bf z}).
$$

Отсюда получаем оценку снизу:
$$
{\bf P}\{\xi^{(t+1)} \in H_j |{\bf z}^{(t)}={\bf z}\} \geq
\sum\limits_{i=0}^d \alpha_{ij}((1-z_{i+1})^s-(1-z_i)^s).
$$
Пусть $Z_{\lambda}=\{{\bf z} \in \R^d: z_i \in
\{0,\frac{1}{{\lambda}},\frac{2}{{\lambda}},\ldots,1\}, z_i\geq z_{i+1}\}$ --
множество векторов популяции из ${\lambda}$ особей. По формуле полной
вероятности получаем безусловную вероятность порождения очередной
особи из~$H_j$ на шаге~$t+1$:
\begin{equation}
\label{basic_geq} {\bf P}\{\xi^{(t+1)} \in H_j\}= \sum
\limits_{{\bf z} \in Z_{\lambda}}{\bf P}\{\xi^{(t+1)} \in H_j|{\bf
z}^{(t)}={\bf z}\}{\bf P}\{{\bf z}^{(t)}={\bf z}\}\geq
\end{equation}
$$
\geq \sum \limits_{{\bf z} \in Z_{\lambda}}\sum\limits_{i=0}^d
\alpha_{ij} ((1-z_{i+1})^s-(1-z_i)^s){\bf P}\{{\bf
z}^{(t)}={\bf z}\}=
$$
$$
=\sum\limits_{i=0}^d
\alpha_{ij}\E[(1-z_{i+1}^{(t)})^s-(1-z_{i}^{(t)})^s].
$$

Ввиду утверждения~\ref{vose} $\E[z_j^{(t+1)}]={\bf P}\{\xi^{(t+1)}
\in H_j\}$. Следовательно, с учетом того, что
$(1-z_{d+1}^{(t)})^s=1$ и $(1-z_{0}^{(t)})^s=0$, из
(\ref{basic_geq}) получаем:
\begin{equation}
\label{simple_expect} \E[z_j^{(t+1)}] \geq
\alpha_{dj}-\sum\limits_{i=1}^d (\alpha_{i,j}-\alpha_{i-1,j})
\E[(1-z_{i}^{(t)})^s].
\end{equation}

Пусть $I_j^+(\A)=\{i \ : \ 1\leq i \leq d, \ \alpha_{i,j}-
\alpha_{i-1,j}\geq 0\}$, $I_j^-(\A)=\{{i \ :} \ {1\leq i \leq d,}
\ \alpha_{i,j}-\alpha_{i-1,j}< 0\}$. Применим к тем слагаемым
последней суммы, где $i \in I_j^-(\A)$, неравенство Иенсена (под
знаком математического ожидания стоит выпуклая функция от
$z_{i}^{(t)}$). Для слагаемых, где $i\in I_j^+(\A)$, можно
воспользоваться оценкой $(1-z_{i}^{(t)})^s \leq 1-z_{i}^{(t)}$. В
результате имеем

\begin{proposition}
Математическое ожидание компонента~$j$ вектора популяции~${\bf
z}^{(t+1)}$ удовлетворяет неравенству
\begin{equation}
\label{thuge} \E[z_j^{(t+1)}] \geq \alpha_{dj}-
\sum\limits_{I_j^+(\A)}
(\alpha_{ij}-\alpha_{i-1,j})(1-\E[z_{i}^{(t)}])-
\sum\limits_{I_j^-(\A)}
(\alpha_{ij}-\alpha_{i-1,j})(1-\E[z_{i}^{(t)}])^s
\end{equation}
для всех $j=1,\ldots,d$.
\end{proposition}

Заметим, что если при всех $i=0,...,d$, $j=1,...,d,$ вероятность
${\bf P}\{{\rm Mut}(\xi)\in H_j\}$ не зависит от выбора $\xi \in
H_i \backslash H_{i+1}$, то всегда можно указать такие матрицы
оценок~$\A$ и $\B$, что $\alpha_{ij} = \beta_{ij}={\bf P}\{{{\rm
Mut}(\xi)\in H_j} | {\xi \in H_i\backslash H_{i+1}}\}$ для всех
$i$ и $j$. Условимся в таком случае называть оператор мутации {\it
ступенчатым} относительно набора линий уровня $\Phi_1,...,\Phi_d$
(или просто ступенчатым, если ясно, о каком наборе подмножеств
идет речь). Матрицу $\M=\A=\B$ с элементами~$\mu_{ij}$,
${i=0,\dots,d,} \ {j=1,\dots,d}$, дающую одновременно верхнюю и
нижнюю оценки для ступенчатого оператора, будем называть матрицей
{\em кумулятивных переходных вероятностей} (по аналогии с матрицей
переходных вероятностей цепи Маркова).

Для получения оценки величины $\E[{\bf z}^{(t)}]$ при
известном векторе средних начальной популяции $\E[{\bf
z}^{(0)}]$ потребуется $t$-кратное применение
неравенства~(\ref{thuge}). С этой целью введем следующее
определение.

Матрицу размерности $d \times d$ с элементами $\mu_{ij}$ будем
называть {\it монотонной}, если $\mu_{i-1,j} \leq \mu_{i,j}$ для
всех $i,j$ от 1 до $d$. Ступенчатый оператор мутации относительно
набора линий уровня $\Phi_1,...,\Phi_d$ будем называть {\em
монотонным} относительно этого набора линий уровня, если его
матрица кумулятивных вероятностей перехода монотонна. Оператор
называется {\em монотонным}, если данное свойство выполнено хотя
бы при одном наборе линий уровня. 


Очевидно, что для любого оператора мутации существуют монотонные
оценки (например, нулевая матрица~$\A$ и матрица~$\B$ со всеми
элементами, равными~1). Однако трудности могут возникать только
из-за отсутствия оценок, достаточно точно передающих свойства
этого оператора.

После несложных преобразований формулы~(\ref{thuge}) получаем

\begin{proposition}
При условии монотонности матрицы $\A$ для всех $j=1,\ldots,d$
имеет место оценка
\begin{equation}
\label{c1} \E[z_j^{(t+1)}] \geq \alpha_{0j}+\sum\limits_{i=1}^d
(\alpha_{ij}-\alpha_{i-1,j})\E[z_{i}^{(t)}].
\end{equation}
\end{proposition}



Ввиду того, что оценка~(\ref{c1}) не зависит от~${\lambda}$ и $s$,
величина~${\bf E}[z_k^{(t)}]$, $k=1,\dots,d, \ t=0,1...$ при
произвольных~${\lambda}$ и $s$ оценивается снизу значением вероятности
${\bf P}\{\xi^{(t)} \in H_k\}$, найденным для ЭА при~${\lambda}=1$, $s=1$,
где оператор мутации имеет кумулятивные переходные вероятности,
совпадающие с монотонными оценками~$\alpha_{ij}$, $i=0,\dots,d, \
j=1,\dots,d$. 

Для получения оценки снизу в достаточно общей ситуации может быть
использовано следующее утверждение. Пусть ${\bf W}$ -- $(d \times
d)$-матрица с элементами $w_{ij}=\alpha_{ij} - \alpha_{i-1,j}$,
$i=1,\dots,d, \ j=1,\dots,d$; ${\bf I}$ -- единичная матрица того
же размера; вектор ${\bf a} = (\alpha_{01},...,\alpha_{0d})$;
$||\cdot||$ -- произвольная матричная норма.

\begin{proposition}
\label{second} Если матрица~$\A$ монотонна и
$||\W^t||\mathop{\longrightarrow}\limits^{t \to \infty} 0$, то для
любого натурального~$t$ имеет место оценка
\begin{equation} \label{c2}
\E[{\bf z}^{(t)}]\geq \E[{\bf z}^{(0)}] {\bf W}^t+{\bf a}({\bf
I}-{\bf W})^{-1}({\bf I}-{\bf W}^t).
\end{equation}
\end{proposition}
{\bf Доказательство.} Рассмотрим в~$\R^d$ последовательность
векторов {\nolinebreak ${\bf u}^{(0)},{\bf u}^{(1)},...,{\bf
u}^{(t)},...$,} где ${\bf u}^{(0)}=\E[{\bf z}^{(0)}],\hspace{2mm}
{\bf u}^{(t+1)}={\bf u}^{(t)}{\bf W}+{\bf a}$. Правая часть
неравенства (\ref{c1}) может только уменьшаться при подстановке
туда нижних оценок компонентов вектора $\E[{\bf z}^{(t)}]$,
поэтому по индукции для любого $t$ получаем: $\E[{\bf z}^{(t)}]
\geq {\bf u}^{(t)}$.

Пользуясь свойствами линейных операторов, ввиду
условия~$||\W^t||\mathop{\longrightarrow}\limits^{t \to \infty} 0$
заключаем, что матрица $({\bf I}-{\bf W})^{-1}$ существует.

Далее по индукции с помощью непосредственной проверки легко
показать, что для любого $t\geq 1 $ имеет место тождество
$$
{\bf u}^{(t)}= {\bf u}^{(0)} {\bf W}^t+{\bf a}({\bf I}-{\bf
W})^{-1}({\bf I}-{\bf W}^t).\vspace{-5mm}
$$
\hfill $\Box$

Правая часть в~(\ref{c2}) с ростом~$t$ приближается к ${\bf
a}({\bf I}-{\bf W})^{-1}$, поэтому предел оценки~(\ref{c2}) не
зависит от распределения популяции~$X^0$. При ступенчатом
операторе мутации и $s=1$ оценка~(\ref{c2}) обращается в
равенство.

Пусть $\alpha_{dj}-\alpha_{0j}<1$, $j=1,\dots,d$. Рассмотрим
матричную норму $||{\bf W}||=\max_{j} \sum_{i=1}^d |w_{ij}|$.
В условиях теоремы $w_{ij}=\alpha_{ij}-\alpha_{i-1,j}\geq 0$,
поэтому $||{\bf W}||=\max_j \sum_{i=1}^d w_{ij}=\max_j
(\alpha_{dj}-\alpha_{0j})<1$, и
условие~$||\W^t||\mathop{\longrightarrow}\limits^{t \to \infty} 0$
выполнено. Заметим, что $\alpha_{dj}-\alpha_{0j}<1$ для всех~$j$,
если при мутации из любого генотипа с ненулевой вероятностью может
быть получен любой наперед заданный генотип.

\subsection{Оценки сверху для случая турнирной селекции} \label{subsec:tourn_up}

По аналогии с выводом формулы~(\ref{basic_geq}) получаем оценку
сверху:
\begin{equation}
\label{simpleabove} \E[z_j^{(t+1)}] \leq
\beta_{dj}-\sum\limits_{i=1}^d
(\beta_{ij}-\beta_{i-1,j})\E[(1-z_{i}^{(t)})^s].
\end{equation}
Отсюда вытекает следующее

\begin{proposition} \label{prop:ub_tourn}
Математическое ожидание вектора популяции удовлетворяет
неравенству
\begin{equation}
\label{thugeabove} \E[z_j^{(t+1)}] \leq
\beta_{dj}-\sum\limits_{I_j^-(\B)}
(\beta_{ij}-\beta_{i-1,j})(1-\E[z_{i}^{(t)}])-
\sum\limits_{I_j^+(\B)}
(\beta_{ij}-\beta_{i-1,j})(1-\E[z_{i}^{(t)}])^s
\end{equation}
для $j=1,\ldots,d$, и если матрица~$\B$ монотонна, то
\begin{equation}
\label{tabove} \E[z_j^{(t+1)}] \leq \beta_{dj}-
\sum\limits_{i=1}^d
(\beta_{ij}-\beta_{i-1,j})(1-\E[z_{i}^{(t)}])^s, \hspace{1em}
j=1,\ldots,d.
\end{equation}
\end{proposition}

Посредством итеративного применения неравенства (\ref{tabove})
можно покомпонентно оценить сверху вектор $\E[{\bf z}^{(t)}]$ для
любого $t$, начав с вектора средних исходной популяции $\E[{\bf
z}^{(0)}]$. Тем не менее, нелинейность правой части (\ref{tabove})
затрудняет получение компактной оценки сверху для произвольного
$t\geq 0$, аналогичной неравенству~(\ref{c2}) из
утверждения~\ref{second}.

Полученные выше оценки не учитывают размер популяции и
справедливы при любом~${\lambda}$. Как показано в~\cite{Er18}, правая часть
формулы~(\ref{tabove}) описывает асимптотику поведения
популяции при
монотонном операторе мутации и ${\lambda} \to \infty$.

\subsection{Верхние оценки для случая $(\mu,\lambda)$-селекции} \label{sec:Model1}

По определению
$(\mu, \lambda)$--селекции, для любого $i=1,\dots,d$ вероятность выбора родительской особи из $H_i$ имеет вид
$$
{\bf P}\{\eta^{1t} \in H_i \ | \ {\bf z}^t={\bf
z}\}=
 \left\{
\begin{array}{ll}
z_i \lambda/ \mu, & \mbox{ если } \ z_i\le\mu/\lambda,\\
1, &  \mbox{ иначе.}
\end{array} \right.
$$
Функцию от $z_i,$ представленную в правой части этого равенства далее будем обозначать через $P_{\rm ch}(z_i)$.

Следующее утверждение доказывается аналогично утверждению~\ref{prop:ub_tourn}.

\begin{proposition} \label{prop:upper_bound}
Если $\B$ монотонна, то для всех $j=1,\ldots,d$ выполнено
\begin{equation}
\label{t_above} \E[z_j^{(t+1)}] \leq \beta_{dj}-
\sum\limits_{i=1}^d (\beta_{ij}-\beta_{i-1,j})\left(1-P_{\rm
ch}(\E[z_{i}^{(t)}]\right).
\end{equation}
\end{proposition}

{\bf Доказательство.}
$$
 \Pr\{\xi^{1,t+1} \in H_j |\z^{(t)}=\z\}=
 $$
 \begin{equation}
\label{t}
=\sum\limits_{i=0}^d \hspace{0.5em}\sum\limits_{\xi \in H_i \backslash H_{i+1}}
\Pr\{{\rm Mut}(\xi) \in H_j \} {\bf P}\{\eta^{1t} = \xi \ | \ {\bf z}^t={\bf
z}\},
\end{equation}
Из выражения (\ref{t}) и определения границ $ \beta_{ij} $
следует для всех $ j = 1, \dots, d $:
$$
\Pr\{\xi^{1,t+1} \in H_j |\z^{(t)}=\z\}\leq \sum\limits_{i=0}^d
\beta_{ij} \sum\limits_{\xi \in H_i \backslash H_{i+1}} \Pr\{ \eta^{1t} =\xi | \z^{(t)}=\z\} =
$$
\begin{equation}\label{eq:conditional}
=\sum\limits_{i=0}^d \beta_{ij} \Pr\{ \xi \in H_i \backslash H_{i+1} | \z^{(t)}=\z\},
\end{equation}
Это дает неравенство:
$$
{\Pr}\{\xi^{1,t+1} \in H_j |{\bf z}^{(t)}={\bf z}\}
 \leq
\sum\limits_{i=0}^d \beta_{ij}(P_{\rm ch}(z_i^{(t)}) - P_{\rm
ch}(z_{i+1}^{(t)})).
$$
По формуле полной вероятности,
\begin{equation}
\label{basic_geq_start} 
\Pr\{\xi^{1,t+1} \in H_j\}= \sum \limits_{\z
\in Z_{\lambda}}\Pr\{\xi^{1,t+1} \in
H_j|\z^{(t)}=\z\}\Pr\{\z^{(t)}=\z\}
\end{equation}
$$
\leq \sum \limits_{\z \in Z_{\lambda}}\sum\limits_{i=0}^d
\beta_{ij} (P_{\rm ch}(z_i^{(t)}) - P_{\rm
ch}(z_{i+1}^{(t)}))\Pr\{\z^{(t)}=\z\}
 =\sum\limits_{i=0}^d
\beta_{ij}\E[P_{\rm ch}(z_i^{(t)}) - P_{\rm
ch}(z_{i+1}^{(t)})]
$$
\begin{equation}
\label{basic_leq} = \beta_{0j} \E[P_{\rm ch}(z_0^{(t)})]- \beta_{dj} \E[P_{\rm ch}(z_{d+1}^{(t)})]
  + \sum\limits_{i=1}^d (\beta_{ij}-\beta_{i-1,j})
\E[P_{\rm ch}(z_{i}^{(t)})],
\end{equation}
где последнее равенство получается перегруппировкой слагаемых. Из предложения~\ref{vose} следует, что
$\E[z_j^{(t+1)}]=\Pr\{\xi^{1,t+1} \in H_j\}$. Следовательно, т.к.
$P_{\rm ch}(z_{d+1}^{(t)})=P_{\rm ch}(0)=0$ и т.к. $P_{\rm ch}(z_{0}^{(t)})=P_{\rm ch}(1)=1$, имеем
$$
\E[z_j^{(t+1)}] \leq \beta_{0j}+\sum\limits_{i=1}^d (\beta_{i,j}-\beta_{i-1,j}) \
\E[P_{\rm ch}(z_i^{(t)})]
=
$$
\begin{equation} \label{basic_leq_end} 
=\beta_{dj}-\sum\limits_{i=1}^d (\beta_{i,j}-\beta_{i-1,j}) \
\E[1-P_{\rm ch}(z_i^{(t)})]. \ \ \ \Box
\end{equation}

Итеративным применением неравенства~(\ref{t_above})
компоненты математического ожидания вектора популяции~$\E[\z^{(t)}] $ могут
быть ограничены до любого~$t $, начиная с исходного
вектора~$\E[\z^{(0)}] $, описывающего популяцию~$\Pi^0.$ В связи с тем, 
что на данный момент замкнутой математической формулы для верхней оценки вектора~$\E[\z^{(t)}] $ не известно, 
далее его оценки получаются описанным алгоритмом. 

\section{Сравнение (1+1)~EA с другими эволюционными алгоритмами}

В данном разделе будет проведено сравнение эволюционной стратегии
$\mbox{(1+1)~EA}$ с другими ЭА. В частности, будет показано, что если
выполняется условие {\em доминирования}, то эта эволюционная
стратегия является наилучшим ЭА с точки зрения вероятности
получения решений требуемого качества к любой заданной итерации. Отсюда будет следовать, что
в условиях доминирования $\mbox{(1+1)~EA}$ является предпочительной
и с точки зрения математического ожидания приспособленности
рекордного решения на любой итерации, а также по математическому
ожиданию числа итераций до получения оптимума.

Для удобства исследования в данном разделе мы будем иначе
определять класс эволюционных алгоритмов, чем в
п.~\ref{sec:ea}. Основное отличие будет состоять в том, что
здесь в явном виде допускается возможность использования
неограниченной памяти о <<предыстории>> работы алгоритма.\footnote{Для того, чтобы подчеркнуть различие 
в характере применения операторов селекции и воспроизведения в ЭА с неограниченной памятью от селекции и
воспроизведения в общей схеме ЭА из п.~\ref{sec:ea} (возможность выбора родителей из всех ранее построенных особей и, как правило, небольшое число входных и выходных генотипов в операторе воспроизведения), здесь будут использоваться обозначения ${\rm
Sel}_{\infty}$ и ${\rm Rep}$ вместо ${\rm Select}$ и ${\rm Reproduce}$.}

Будем называть {\em эволюционным алгоритмом с
неограниченной памятью} следующий рандомизированный
алгоритм: начальная популяция $\xi^{1,0},\dots,\xi^{{\lambda},0}$
строится некоторым случайным или детерминированным
способом. Пусть~$t$, как и ранее, обозначает номер
итерации. Последовательность генотипов, построенных в ЭА с
неограниченной памятью к началу итерации~$t$, обозначим
через $\sigma^{t-1}$, а множество всех входящих в нее
элементов -- через~${\rm A}^{(t-1)}$. На каждой
итерации~$t, \ t=1,2,\dots,$ вычисляются ${\lambda}''$ генотипов
потомков $\xi^{1,t},\dots,\xi^{{\lambda}''t}$ посредством
применения оператора воспроизведения ${\rm
Rep}(\eta^1,\dots,\eta^{{\lambda}'})$, где $\eta^1,\dots,\eta^{{\lambda}'}$
-- некоторые из ранее построенных особей. Родительские
особи $\eta^1,\dots,\eta^{{\lambda}'}$ на итерации~$t$ выбираются с
помощью рандомизированного оператора {\em селекции с
неограниченной памятью} ${\rm Sel}_{\infty}:{\mathcal
B}^{{\lambda}+{\lambda}''(t-1)} \to {\mathcal B}^{{\lambda}'}$, так что ${{\rm
Sel}_{\infty}(\sigma^{t-1})} \subseteq {\rm A}^{(t-1)}.$

Работа ЭА с неограниченной памятью представляет собой
итерации следующей композиции случайных отображений:
$$
X^{t}={\rm Rep}({\rm Sel}_{\infty}(X^0,\dots,X^{t-1})), \ \
t=1,2,\dots.
$$
\noindent Алгоритм, удовлетворяющий этому определению, обозначим
через~EA.

Начнем со следующего определения из теории вероятностей.

\begin{definition} Случайная величина~$Y$
<<стохастически больше>> случайной величины~$Y'$, если
\begin{equation} \label{stochgreater}
P\{Y \ge \phi\} \ge P\{Y' \ge \phi\}
\end{equation}
при любом~$\phi\in \R$.
\end{definition}

Зачастую (см., например,~\cite{ZP82}) в определении
<<стохастически большей>> случайной величины
вместо~(\ref{stochgreater}) требуют другого (аналогичного)
неравенства
\begin{equation} \label{stochgreater1}
P\{Y > \phi\} \ge P\{Y' > \phi\},
\end{equation}
однако для совместимости с результатами из~\cite{BE06} мы будем
использовать неравенство~(\ref{stochgreater}).


Пусть текущий генотип на итерации~$t, \ t=0,1,\dots,$ эволюционной
стратегии $\mbox{(1+1)~EA}$ обозначается через~$\zeta^{(t)}$.

Будем обозначать максимум функции приспособленности на
какой-либо последовательности
генотипов~$\sigma=(\xi_1,\dots,\xi_k)$
через~$\check\Phi(\sigma)$, \ie
$\check\Phi(\sigma)=\max_{i=1,\dots,k} \Phi(\xi_i)$.

Сравнение оператора воспроизведения из алгоритма~EA и оператора
мутации эволюционной стратегии~$\mbox{(1+1)~EA}$ будет
осуществляться на основе следующего определения.

\begin{definition} \label{definedom}
Оператор воспроизведения~${\rm Rep}$ {\it доминируется}
оператором мутации~${\rm Mut}$, если для любой
${\lambda}'$-элементной последовательности
генотипов~$X'=(\xi^1,\dots,\xi^{{\lambda}'})$ и любого~$\eta \in
{ B}$, таких что $\Phi(\eta)\geq \check\Phi(X'),$
выполняется следующее условие при всех~$\phi >\Phi(\eta)$:
\begin{equation}\label{DefDom}
{\bf P}\left\{\Phi({\rm Mut}(\eta))\geq \phi \right\}\geq {\bf
P}\left\{\check\Phi({\rm Rep}(X'))\geq \phi \right\}.
\mbox{\hspace{1em}}
\end{equation}
\end{definition}

Данное определение может интерпретироваться как требование того,
чтобы вероятность получения генотипа достаточно высокой
приспособленности посредством оператора мутации~${\rm Mut}$ была
бы не меньше, чем вероятность такого же события для оператора
рекомбинации~${\rm Rep}$, если входной генотип оператора~${\rm
Mut}$ не уступает ни одному из родительских генотипов на
входе~${\rm Rep}$.

В качестве иллюстративного примера доминирования рассмотрим задачу
максимизации приспособленности $\Phi(\xi)\equiv ONEMAX(\xi)\equiv
\xi_1+\dots+\xi_n$, где $X=B=\{0,1\}^n$, $l=n$, когда~${\rm
Rep}$ -- оператор равномерного кроссинговера с одним
потомком (второй потомок не строится). Пусть мутация~${\rm Mut}$
устроена специальным образом: гены со значением~1 сохраняются без
изменений, в то время как биты со значением 0 изменяются с
вероятностью~1/2. Очевидно, такой оператор~${\rm Mut}$
доминирует~${\rm Rep}$: рассмотрим лучшего из двух
родителей. Шансы на мутацию в любом гене со значением~0 не ниже
шансов на улучшение гена со значением 0 при действии
рассматриваемого кроссинговера.


Следующая теорема показывает, что если начальный генотип
$\mbox{(1+1)~EA}$ не уступает по приспособленности лучшей
особи начальной популяции EA, то на любой итерации
распределение вероятностей текущего генотипа
$\mbox{(1+1)~EA}$ также не уступает распределению лучшего
из найденных генотипов в EA тогда и только тогда, когда~Mut
доминирует~Rep.

\begin{theorem}
\label{EAdominate} Для того чтобы к любому алгоритму EA из класса
ЭА с неограниченной памятью при
\begin{equation}
\label{EAstart} \Phi(\zeta^{(0)}) \geq \max\limits_{i=1,\dots,{\lambda}}
\Phi(\xi^{(i,0)})
\end{equation}
были применимы неравенства
\begin{equation}
\label{EAclaim} {\bf P}\{\Phi(\zeta^{(t)})\geq \phi\} \geq {\bf
P}\{\check{\Phi}( \sigma^{t})\geq \phi\},  \ \ t\in {\mathbb
Z}_{+}, \ \ \phi\in \R,
\end{equation}
необходимо и достаточно, чтобы оператор~${\rm Rep}$ доминировался
оператором ${\rm Mut}$.
\end{theorem}

{\bf Доказательство.}

1.  Достаточность. Пусть множество всевозможных значений
функции приспособленности есть $\{\Phi_0,\dots,\Phi_d\},$ где
$\Phi_0<\dots<\Phi_d$. Пользуясь индукцией по~$t$, предположим,
что
$$
{\bf P}\{\Phi(\zeta^{(t-1)})\geq \Phi_j\} \geq {\bf
P}\{\check{\Phi}(\sigma^{t-1})\geq \Phi_j\} \mbox{\rm  для всех
} j=0,\dots,d
$$
(базис индукции при~$t=0$ следует из формулировки теоремы). На
итерации~$t$ алгоритма~EA имеем
$\check{\Phi}(\sigma^{t})=\max\{\Phi(\xi'),
\check{\Phi}(\sigma^{t-1})\}$, где~$\xi'$ -- генотип с
наибольшей приспособленностью среди потомков, построенных
оператором ${\rm Rep}(\eta^1,\dots,\eta^{{\lambda}'})$, при
$\{\eta^1,\dots,\eta^{{\lambda}'}\} \subseteq {\rm A}^{(t-1)}$,
выбранных оператором~${\rm Sel}_{\infty}$.

Зафиксируем произвольное~$j, \ $ $j \in \{0,\dots, d\}$, и введем
обозначения для каждого $i=0,\dots,j$:
$$
 p_i={\bf P}\{\check{\Phi}(\sigma^{t-1}) \geq \Phi_i\}, \ \
 v_{ij}={\bf P}\{\Phi(\xi') \geq \Phi_j \ | \ \
 \check{\Phi}(\sigma^{t-1}) =  \Phi_i \}.
$$
Тогда
\[
{\bf P}\{\check{\Phi}(\sigma^{t})\geq \Phi_j\}= p_j+ {\bf
P}\{\Phi(\xi') \geq \Phi_j , \check{\Phi}(\sigma^{t-1}) < \Phi_j\}
=
\]
\begin{equation} \label{totalEA}
= p_j+\sum\limits_{i=0}^{j-1} v_{ij}(p_i-p_{i+1})
\end{equation}
по формуле полной вероятности.

Аналогичные рассуждения для $\mbox{(1+1)~EA}$ дают
\begin{equation} \label{total1plus1}
{\bf P}\{\Phi(\zeta^{(t)})\geq
\Phi_j\}=q_j+\sum\limits_{i=0}^{j-1} w_{ij}(q_i-q_{i+1}),
\end{equation}
где $q_i={\bf P}\{\Phi(\zeta^{(t-1)}) \geq \Phi_i\}$ и
$w_{ij}={\bf P}\{\Phi(\zeta')\geq \Phi_j \ | \ \Phi(\zeta^{(t-1)})
= \Phi_i\}$, \linebreak $i=0,\dots,j$, при $\zeta'={\rm
Mut}(\zeta^{(t-1)})$.

Условие доминирования означает, что для любых $i,k, \ i\leq k, \ k
< j$, выполняется $v_{ij} \leq w_{kj}$. Пусть
$\hat{v}_{ij}=\max\limits_{l=0,\dots,i} v_{lj}$ для каждого
$i=0,\dots,j-1$. Тогда $\hat{v}_{i-1,j}\leq \hat{v}_{i,j}$ и
$$
v_{ij}\leq \hat{v}_{ij} \leq w_{ij}, \ \ i=0,\dots,j-1.
$$
Используя эти свойства, выражения~(\ref{totalEA}) и
(\ref{total1plus1}),
получаем:
$$
{\bf P}\{\check{\Phi}(\sigma^{t})\geq \Phi_j\}=p_j +
\sum\limits_{i=0}^{j-1} v_{ij}(p_i-p_{i+1}) \leq p_j +
\sum\limits_{i=0}^{j-1} \hat{v}_{ij}(p_i-p_{i+1})=
$$
$$
=\hat{v}_{0j}+\sum\limits_{i=1}^{j-1}(\hat{v}_{ij}-\hat{v}_{i-1,j})p_i+
(1-\hat{v}_{j-1,j})p_j \leq
$$
$$
\leq
\hat{v}_{0j}+\sum\limits_{i=1}^{j-1}(\hat{v}_{ij}-\hat{v}_{i-1,j})q_i+
(1-\hat{v}_{j-1,j})q_j =
$$
$$
= q_j + \sum\limits_{i=0}^{j-1} \hat{v}_{ij}(q_i-q_{i+1})\leq q_j
+ \sum\limits_{i=0}^{j-1} w_{ij}(q_i-q_{i+1})= {\bf
P}\{\Phi(\zeta^{(t)})\geq \Phi_j\}.\ \ \
$$
\\

2. Необходимость. Рассмотрим алгоритм EA, где ${\lambda}={\lambda}'$, а
оператор селекции таков, что всякий раз выбираются ${\lambda}'$ последних
построенных EA генотипов из последовательности~$\sigma^t$.
Согласно условиям теоремы при любой ${\lambda}'$-элементной
последовательности
генотипов~$X^0=(\xi^{(1,0)},\dots,\xi^{({\lambda}',0)})$ и
любом~$\zeta^{(0)} \in { B}$, удовлетворяющих
условию~(\ref{EAstart}), \ie таких, что $\Phi(\zeta^{(0)})\geq
\check\Phi(X^0)$, для всех~$\phi > \Phi(\zeta^{(0)})$ имеем:
$$
{\bf P}\left\{\Phi({\rm Mut}(\zeta^{(0)}))\geq \phi \right\} =
{\bf P}\{\Phi(\zeta^{(1)})\geq \phi\}\geq
$$
$$
\geq {\bf P}\{\check{\Phi}( \sigma^{1})\geq \phi\}={\bf
P}\left\{\check\Phi({\rm Rep}(X^0))\geq \phi \right\},
$$
\ie оператор~${\rm Rep}$ доминируется оператором ${\rm
Mut}$. $\Box$


В \cite{BE06} было показано, что при $p_{\rm m} \le 0.5,$ оператор мутации из КГА для задачи ONEMAX
доминирует сам себя, т.е. является {\em монотонным}. Аналогичный результат
получен для ЗВП на графах специальной структуры. Большинство
известных операторов мутации не имеют свойства монотонности при решении
нетривиальных задач. В частности, известно, что при наличии локальных, но
не глобальных, оптимумов в смысле метрики Хэмминга, оператор мутации из
КГА не будет монотонным.~\cite{BE06}

\subsection{Монотонные операторы воспроизведения}

В некоторых случаях для оператора~${\rm Rep}$ существует
достаточно простой способ построения оператора мутации,
доминирующего его. Определим оператор мутации, соответствующий
оператору~${\rm Rep}$ следующим образом
$$
{\rm Mut}_{\rm Rep}(\zeta)=\mbox{\rm  argmax
}(\Phi(\zeta),\Phi(\xi^1), \dots,\Phi(\xi^{{\lambda}''})),
$$
где $(\xi^1,\dots,\xi^{{\lambda}''})={\rm Rep}(\zeta,\dots,\zeta)$. Таким
образом, в ${\rm Mut}_{\rm Rep}$ сначала оператор
воспроизведения~${\rm Rep}$ применяется к набору идентичных
родительских генотипов, и после этого результат выбирается как
генотип с наибольшей приспособленностью среди генотипов, имеющихся
на входе и выходе~${\rm Rep}$. В случае, если при этом обнаружится
несколько генотипов с одинаковой приспособленностью, будем
предполагать, что ${\rm Mut}_{\rm Rep}(\zeta)$ выбирается среди
них равновероятно. 

\begin{definition} \label{DefMonot}
Оператор воспроизведения ${\rm Rep}$ называется {\it
монотонным}, если для любых двух последовательностей из ${\lambda}'$
генотипов $\xi^1,\dots,\xi^{{\lambda}'}$ и
$\bar{\xi}^1,\dots,\bar{\xi}^{{\lambda}'}$, таких что
\begin{equation}\label{DefMon0}
\Phi(\xi^1)\leq\Phi(\bar{\xi}^1),\dots,\Phi(\xi^{{\lambda}'})\leq\Phi(\bar{\xi}^{{\lambda}'}),
\end{equation}
следующие условия выполняются для всех~$\phi$:
\begin{equation} \label{DefMon1}
{\bf P}\left\{\max\limits_{i=1,\dots,{\lambda}''} \Phi(\bar{\eta}^i)\geq
\phi \right\}\ge
 {\bf P}\left\{\max\limits_{i=1,\dots,{\lambda}''} \Phi(\eta^i)\geq \phi
\right\}, \mbox{\hspace{1em}}
\end{equation}
где $(\eta^1,\dots,\eta^{{\lambda}''})={\rm Rep} (\xi^1,\dots,\xi^{{\lambda}'})$ и
$(\bar{\eta}^1,\dots,\bar{\eta}^{{\lambda}''})={\rm Rep}
(\bar{\xi}^1,\dots,\bar{\xi}^{{\lambda}'})$.
\end{definition}

Условие монотонности оператора воспроизведения означает, что
замена родительских генотипов на генотипы большей или равной
приспособленности не приводит к снижению шансов получения
достаточно приспособленных потомков на его выходе.

Заметим, что если условия~(\ref{DefMon0}) выполнены как равенства
для родительских генотипов~$\xi^1,\dots,\xi^{{\lambda}'}$ и
$\bar{\xi}^1,\dots,\bar{\xi}^{{\lambda}'}$, то ввиду~(\ref{DefMon1})
распределения вероятностей приспособленности лучшего потомка для
${\rm Rep}(\xi^1,\dots,\xi^{{\lambda}'})$ и ${\rm
Rep}(\bar{\xi}^1,\dots,\bar{\xi}^{{\lambda}'})$ должны совпадать.

Если оператор ${\rm Rep}$ является монотонным, то ${\rm
Mut}_{\rm Rep}$ доминирует ${\rm Rep}$ по построению, в
таком случае алгоритм EA с оператором воспроизведения~${\rm
Rep}$ может сравниваться по теореме~\ref{EAdominate} с
$\mbox{(1+1)~EA}$, где используется оператор мутации ${\rm
Mut}_{\rm Rep}$. Следующее следствие непосредственно
вытекает из теоремы~\ref{EAdominate}.

\begin{corollary}
\label{EAcompare} Пусть в EA используется  монотонный оператор
воспроизведения~${\rm Rep}$, и оператор ${\rm Mut}_{\rm Rep}$
используется в $\mbox{(1+1)~EA}$. Пусть, кроме того,
$\mbox{(1+1)~EA}$ начинает работу с генотипа~$\zeta^{(0)}$,
такого что $\Phi(\zeta^{(0)}) \geq \max\limits_{i=1,\dots,{\lambda}}
\Phi(\xi^{(i,0)})$. Тогда для всех $t\geq 0$ и всех $\phi\in \R$
выполняется
\begin{equation}
\label{EAclaim1} {\bf P}\{\Phi(\zeta^{(t)})\geq \phi\} \ge {\bf
P}\{\check{\Phi}(\sigma^t)\geq \phi\}.
\end{equation}
\end{corollary}

Частным случаем монотонного оператора воспроизведения при
${\lambda}'={\lambda}''=1$ является монотонный оператор мутации.
Непосредственно из определений следует, что монотонный оператор
мутации доминирует сам себя, и по следствию~\ref{EAcompare} в
таком случае $\mbox{(1+1)~EA}$ является <<наилучшим>> методом в
классе эволюционных алгоритмов с неограниченной памятью. 

Приведем простой пример монотонного оператора воспроизведения, где
${\lambda}'=2,{\lambda}''=1$. Пусть $\Phi \equiv {\rm ONEMAX}$ и оператор $\rm
Rep$ построен на основе равномерного кроссинговера, однако перед
применением равномерного кроссинговера к генам одного из двух
родительских генотипов применяется перестановка, выбранная
равновероятно. В качестве результата воспроизведения используется
первый из двух генотипов, полученных в результате равномерного
кроссинговера.

Заметим, что без использования случайной перестановки такой
оператор не был бы монотонным (например, для функции
$\Phi(\xi) = \sum_{i=1}^l \xi_i$ условие~(\ref{DefMon1})
нарушается на родительских генотипах~$\xi^1=(1,0),
\xi^2=(0,1)$ $\bar\xi^1=(1,0), \bar\xi^2=(1,0)$
при~$\phi=2$).

Все упомянутые выше примеры относятся к задачам регулярной
структуры. На практике, однако, задачи комбинаторной оптимизации,
как правило, не имеют такой структуры и свойство монотонности
оператора воспроизведения выполняется редко. В частности, при
наличии локальных оптимумов функции приспособленности, не
являющихся глобальными, оператор точечной мутации не может быть монотонным. 

\subsection{Среднее время достижения оптимума и средняя
приспособленность на заданной итерации}

Рассмотрим среднее число обращений к оператору воспроизведения до
достижения множества генотипов требуемого качества (при этом наибольший интерес
представляет множество генотипов, кодирующих
оптимальные решения). Вместе с тем,
будем рассматривать и математическое ожидание приспособленности
рекордного решения, получаемого к заданной итерации ЭА с
неограниченной памятью.

Обозначим через~$t^{(1+1)}_{\phi}$ среднее число итераций
$\mbox{(1+1)~EA}$ до получения генотипа с
приспособленностью не ниже~$\phi$. Среднее число итераций
ЭА с неограниченной памятью до получения генотипа с
приспособленностью не ниже~$\phi$ обозначим
через~$t^{EA}_{\phi}$.

\begin{corollary}
\label{hitcompare} Пусть оператор~${\rm Rep}$ доминируется
оператором~${\rm Mut}$ и
\begin{equation}
\label{EAstart_corrol} \Phi(\zeta^{(0)}) \geq
\max\limits_{i=1,\dots,{\lambda}} \Phi(\xi^{(i,0)}),
\end{equation}
тогда: \\
(i) всех $t\ge 0$ выполняется
$$
{\bf E}[\Phi(\zeta^{(t)})] \ge {\bf
E}[\check{\Phi}(\sigma^t)];
$$
(ii) если $t^{EA}_{\phi}$ конечно, то $t^{EA}_{\phi} \geq t^{(1+1)}_{\phi}$.\\
\end{corollary}
{\bf Доказательство.} (i)
Пусть $F_1(\phi)$ и $F_2(\phi)$ обозначают функции распределения
случайных величин~$\Phi(\zeta^{(t)})$ и $\check{\Phi}(\sigma^t)$
соответственно. Тогда из теоремы~\ref{EAdominate} следует
неравенство $F_{1}(\phi) \le F_{2}(\phi)$ для любого $\phi \in
\R$. Таким образом, по свойствам математического ожидания (см.,
например,~\cite{Borovk,Gnedenko})
\begin{equation}\label{gned}
{\bf E}[\Phi(\zeta^{(t)})]= \int_0^{\infty} (1-F_1(\phi))
d\phi \ge \int_0^{\infty} (1-F_2(\phi)) d\phi = {\bf
E}[\check{\Phi}(\sigma^t)].
\end{equation}

(ii) 
Аналогично неравенству~(\ref{gned}) при конечном ${\bf E}[t]$
имеем
$$
t^{EA}_{\phi}= \sum_{k=0}^\infty (1-{\bf
P}\{\check{\Phi}(\sigma^k)\geq\phi\})
$$
и
$$
t^{(1+1)}_{\phi}=\sum_{k=0}^\infty (1-{\bf
P}\{\Phi(\zeta^{(k)})\geq \phi\}).
$$
Применение теоремы~\ref{EAdominate} завершает доказательство.
$\Box$

Теорема~\ref{EAdominate} и следствие~\ref{hitcompare} показывают,
что если требуется сделать выбор между ЭА и эволюционной
стратегией $\mbox{(1+1)~EA}$, то в условиях применимости этих
результатов предпочтение следует отдать $\mbox{(1+1)~EA}$.

\section*{Приложение 1. Список задач}

\begin{exercise}
Показать, что при известных значениях приспособленности особей текущей
популяции~$\Pi^t$ селекция всех родительских особей для построения новой
популяции~$\Pi^{t+1}$ может быть выполнена в КГА за время~$O({\lambda}\log_2({\lambda}))$.
\end{exercise}

\begin{exercise}
Предложить взаимно-однозначное представление решений задачи о максимальном
разрезе в графе при~${l=n-1}$.
\end{exercise}


\begin{exercise}Показать, что для задачи о наименьшем
вершинном покрытии при $v_j \in C \Leftrightarrow \xi_j=1$ задача
оптимальной рекомбинации эффективно разрешима.
\end{exercise}

\begin{exercise}Описать алгоритм, осуществляющий мутацию
<<2-замена>> в кодировке решений задачи коммивояжера с помощью
перестановок.
\end{exercise}

\end{document}